\documentclass[smallextended,natbib]{svjour3}

\usepackage[english]{babel}
\usepackage{amssymb,amsmath}
\usepackage{graphicx,subfigure}
\usepackage{mathptmx}
\usepackage{qtree,linguex}
\usepackage{float}
\usepackage{subfigure}
\usepackage{colordvi}

\newcommand{\pref}[1]{(\ref{#1})}
\newcommand{\Sec}[1]{Sec.~\ref{#1}}
\newcommand{\Eq}[1]{Eq.~(\ref{#1})}
\newcommand{\Fig}[1]{Fig.~\ref{#1}}
\newcommand{\Def}[1]{Def.~\ref{#1}}
\newcommand{\Tab}[1]{Tab.~\ref{#1}}

\newcommand{\rank}{\mathrm{rank}}
\newcommand{\id}{\mathrm{id}}

\newcommand{\lab}{\mathrm{label}}
\newcommand{\merge}{\mathrm{merge}}
\newcommand{\move}{\mathrm{move}}
\newcommand{\head}{\mathrm{head}}
\newcommand{\first}{\mathrm{first}}
\newcommand{\shift}{\mathrm{shift}}
\newcommand{\feat}{\mathrm{feat}}
\newcommand{\leav}{\mathrm{leaves}}
\newcommand{\rep}{\mathrm{replace}}
\newcommand{\cons}{\mathrm{cons}}
\newcommand{\ext}{\mathrm{ex}}
\newcommand{\sel}{\mathrm{sel}}
\newcommand{\lic}{\mathrm{lic}}

\newcommand{\hmerge}{\mathrm{hmerge}}
\newcommand{\hmove}{\mathrm{hmove}}

\newcommand{\bflab}{\mathbf{label}}
\newcommand{\bfmerge}{\mathbf{merge}}
\newcommand{\bfmove}{\mathbf{move}}
\newcommand{\bfhead}{\mathbf{head}}
\newcommand{\bffirst}{\mathbf{first}}
\newcommand{\bfshift}{\mathbf{shift}}
\newcommand{\bffeat}{\mathbf{feat}}
\newcommand{\bfleav}{\mathbf{leaves}}
\newcommand{\bfrep}{\mathbf{replace}}
\newcommand{\bfcons}{\mathbf{cons}}
\newcommand{\bfext}{\mathbf{ex}}
\newcommand{\bfmax}{\mathbf{max}}
\newcommand{\bfsel}{\mathbf{sel}}
\newcommand{\bflic}{\mathbf{lic}}
\newcommand{\bfubfeat}{\mathbf{ubfeat}}

\newcommand{\Lex}{\mathrm{Lex}}
\newcommand{\DomMerge}{\mathrm{Dom}_\mathrm{merge}}
\newcommand{\DomMove}{\mathrm{Dom}_\mathrm{move}}

\newcommand{\bfDomMerge}{\mathrm{Dom}_\mathbf{merge}}
\newcommand{\bfDomMove}{\mathrm{Dom}_\mathbf{move}}

\journalname{Journal of Logic, Language and Information}

\bibpunct{(}{)}{,}{}{}{;}

\begin{document}

%---------------------------------------------title page--------------------------------------------

\title{Geometric representations for minimalist grammars}
\titlerunning{Geometric representations for minimalist grammars}

\author{Peter beim Graben
    \and
    Sabrina Gerth}

\authorrunning{beim Graben \& Gerth}

\institute{
    Peter beim Graben \at
    Dept. of German Language and Linguistics, and \\
    Bernstein Center for Computational Neuroscience \\
    Humboldt-Universit\"at zu Berlin \\
    Unter den Linden 6 \\
    D-10099 Berlin \\
    \email{peter.beim.graben@hu-berlin.de}
      \and
    Sabrina Gerth \at
    Department Linguistics, \\
    University of Potsdam, Germany
    }

\date{\today}
\maketitle

\begin{abstract}
We reformulate minimalist grammars as partial functions on term algebras for strings and trees. Using filler/role bindings and tensor product representations, we construct homomorphisms for these data structures into geometric vector spaces. We prove that the structure-building functions as well as simple processors for minimalist languages can be realized by piecewise linear operators in representation space. We also propose harmony, i.e. the distance of an intermediate processing step from the final well-formed state in representation space, as a measure of processing complexity. Finally, we illustrate our findings by means of two particular arithmetic and fractal representations.
\end{abstract}


% ------------------------------------- Section -----------------------------------
\section{Introduction}
\label{sec:intro}

Geometric approaches to cognition in general and to symbolic computation in particular became increasingly popular during the last two decades. They comprise conceptual spaces for sensory representations \citep{Gaerdenfors04}, latent semantic analysis for the meanings of nouns and verbs \citep{CederbergWiddows03}, and tensor product representations for compositional semantics \citep{Blutner09, Aerts09}. According to the dynamical system approach to cognition \citep{Gelder98, GrabenPotthast09a}, mental states and their temporal evolution are represented as states and trajectories in a dynamical system's state space. This approach has been used, e.g., for modeling logical inferences \citep{BalkeniusGaerdenfors91, Mizraji92} and language processes \citep{GrabenGerthVasishth08, Tabor09}. Interpreting the states of a dynamical system as activation vectors of neural networks, includes also connectionist approaches of cognitive modeling into geometric cognition \citep{GerthGraben09b, Huyck09, VosseKempen09}.

One particularly significant contribution in this direction is Smolensky's Integrated Connectionist/Symbolic Architecture (ICS) \citep{Smolensky06, SmolenskyLegendre06a}. This is a dual-aspect approach where subsymbolic dynamics of neural activation patterns at a lower-level description become interpreted as symbolic cognitive computations at a higher-level description by means of filler/role bindings through tensor product representations. Closely related to ICS is dynamic cognitive modeling (DCM) \citep{GrabenPotthast09a, GrabenPotthastTA}, which is a top-down approach for the construction of neurodynamical systems from symbolic representations in continuous time.

So far, ICS/DCM architectures have been successfully employed for phonological  \citep{Smolensky06, SmolenskyLegendre06a} and syntactic computations \citep{Smolensky06, SmolenskyLegendre06a, GrabenGerthVasishth08} in the fields of computational linguistics and computational psycholinguistics using mainly context-free grammars and appropriate push-down automata \citep{HopcroftUllman79}. However, as natural languages are known to belong to the complexity class of mildly context-sensitive languages within the Chomsky hierarchy \citep{Shieber85, Stabler04}, more sophisticated formal grammars have been developed, including tree-adjoining grammars (TAG) \citep{JoshiLevyTakahashi75}, multiple context-free grammars \citep{SekiEA91} and minimalist grammars \citep{Stabler97, StablerKeenan03}. In particular, Stabler's formalism of minimalist grammars (MG) codifies most concepts of generative linguistics (e.g. from Government and Binding Theory \citep{Chomsky81, Haegeman94} and Chomsky's Minimalist Program \citep{Chomsky95, Weinberg01}) in a mathematically rigorous manner. In early MG this has been achieved by defining minimalist trees and the necessary transformations by means of set and graph theoretic operations. In its later development, minimalist trees have been abandoned in favor of chain-based calculus \citep{Harkema01, StablerKeenan03} due to Harkema's statement that ``the geometry of a [minimalist] tree is a derivational artifact of no relevance [...]'' \citep[p.~82]{Harkema01}. Based on these results MG could be recast into multiple context-free grammars for further investigation \citep{Michaelis01, Harkema01}.

Recently, \citet{Gerth06} and \citet{GerthGraben09b} revived the ideas of early MG, by presenting two ICS/DCM studies for the processing of minimalist grammars in geometric representation spaces, because minimalist tree representations could be of significance for psycholinguistic processing performance. In these studies, different filler/role bindings for minimalist feature arrays and minimalist trees have been used: one purely arithmetic representation for filler features and syntactic roles \citep{Gerth06}; and another one, combining arithmetic and numerical representations into a fractal tensor product representation \citep{GerthGraben09b}. Until now, these studies lack proper theoretical justification by means of rigorous mathematical treatment. The present work aims at delivering the required proofs. Moreover, based on the metric properties of representation space, we present an extension of MG toward harmonic MG, for providing a complexity measure of minimalist trees that might be of relevance for psycholinguistics.

The paper is structured as follows. In \Sec{sec:mg} we algebraically recapitulate Stabler's original proposals for minimalist grammars which is required for subsequent dynamic cognitive modeling. We also illustrate the abstract theory by means of a particular linguistic example in \Sec{sec:applMG}. Next, we build an ICS/DCM architecture in \Sec{sec:isc} by mapping filler/role decompositions of minimalist data structures onto tensor product representations in geometric spaces. The main results of the section are summarized in two theorems about minimalist representation theory. We also introduce harmonic minimalist grammar (HMG) here, by proposing a harmony metric for minimalist trees in representation space. In \Sec{sec:appl} we resume the linguistic example from \Sec{sec:applMG} and construct arithmetic and fractal tensor product representations for our minimalist toy-grammar. The paper concludes with a discussion in \Sec{sec:disc}.

% ------------------------------------- Section -----------------------------------
\section{Minimalist Grammars Revisited}
\label{sec:mg}

In this section we rephrase derivational minimalism \citep{Stabler97, StablerKeenan03, Michaelis01} in terms of term algebras \citep{Kracht03} for feature strings and trees which is an important prerequisite for the aim of this study, namely vector space representation theory. Moreover, following \citet{Harkema01}, we disregard the original distinction between ``strong'' and ``weak'' minimalist features that allow for ``overt'' vs. ``covert'' movement and for merge with or without head adjunction, respectively. For the sake of simplicity we adopt the notations of ``strict minimalism'' \citep{Stabler99, Michaelis04}, yet not taking its more restricted move operation, the specifier island condition \citep{GartnerMichaelis07a}, into account.

% ------------------------------------- Section -----------------------------------
\subsection{Feature strings}
\label{sec:featstr}

Consider a finite set of \emph{features} $F_F$ and its Kleene closure $F_F^*$. The elements of $F_F^*$ can be regarded as terms over the signature $F_F \cup \{ \varepsilon \}$, where the empty word $\varepsilon$ has arity 0 and features $f \in F_F$ are unary function symbols. Then, the term algebra $T_F$ is inductively defined through (1) $\varepsilon \in T_F$ is a term. (2) If $s \in T_F$ is a term and $f \in F_F$, then $f(s) \in T_F$. Thus, a string (or likewise, an array) $s = f_1 f_2 \ldots f_p \in F_F^*$, $p \in \mathbb{N}_0$, of features $f_i \in F_F$ is regarded as a term $s = (f_1 \circ f_2 \circ \ldots \circ f_p)(\varepsilon) = f_1(f_2(\ldots(f_p(\varepsilon)))) \in  T_F$, where ``$\circ$'' denotes functional composition (the set $\mathbb{N}_0$ contains the non-negative integers $0, 1, 2, \dots$).

First, we define two string functions for preparing the subsequent introduction of minimalist grammars.
\begin{definition}\label{def:strfnc}
Let $s \in T_F$ be a feature string with $s = f(r)$, $f \in F_F$, $r \in T_F$.
\begin{enumerate}
\item The \emph{first feature} of $s$ is obtained by the function $\first: T_F \setminus \{ \varepsilon \} \to F_F$, $\first(s) = \first(f(r)) = f$.
\item In analogy to the \emph{left-shift} in symbolic dynamics \citep{LindMarcus95} we define $\shift: T_F \setminus \{ \varepsilon \} \to T_F$, $\shift(s) = \shift(f(r)) = r$.
\end{enumerate}
\end{definition}
Basically, the functions $\first$ and $\shift$ correspond to the LISP functions car and cdr, respectively.

% ------------------------------------- Section -----------------------------------
\subsection{Labeled trees}
\label{sec:labtree}

In early MG, a minimalist expression is a finite, binary, and ordered tree endowed with the relation of (immediate) projection among siblings and with a labeling function mapping leaves onto feature strings \citep{Stabler97, Michaelis01}. Such trees become terms from a suitably constructed term algebra $T_A$, as follows. As signature of $T_A$ we choose the ranked alphabet $A = T_F \cup \{ \mathtt{<},  \mathtt{>} \}$, where $T_F$ is the previously introduced algebra of feature strings, and $\rank_A : A \to \mathbb{N}_0$. Feature strings are ranked as constants through $\rank_A(s) = 0$ for all $s \in T_F$. Furthermore, the minimalist projection indicators, $\mathtt{<},  \mathtt{>}$, are regarded as binary function symbols: $\rank_A(\mathtt{<}) = \rank_A(\mathtt{>})  = 2$. Then we define by means of induction: (1) Every $s \in T_F$ is a term,  $s \in T_A$. (2) For terms $t_0, t_1 \in T_A$, $\mathtt{<}(t_0, t_1) \in T_A$ and $\mathtt{>}(t_0, t_1) \in T_A$. Then, $\mathtt{<}(t_0, t_1)$ denotes a minimalist tree with root $\mathtt{<}$, left subtree $t_0$ and right subtree $t_1$. The root label $\mathtt{<}$ indicates that $t_0$ ``projects over'' $t_1$. By contrast, in the tree $\mathtt{>}(t_0, t_1)$ $t_1$ ``projects over'' $t_0$.
\begin{definition}\label{def:simpcomp}
A minimalist tree $t \in T_A$ is called \emph{complex} if there are terms $t_0, t_1 \in T_A$ and $f = \mathtt{>}$ or $f = \mathtt{<}$ such that $t = f(t_0, t_1)$. A tree that is not complex is called \emph{simple}.
\end{definition}

In correspondence to \citet{SmolenskyLegendre06a} and \citet{Smolensky06}, we define the following functions for handling minimalist trees.
\begin{definition}\label{def:treefnc1}
Let $t \in T_A$ be given as $t = f(t_0, t_1)$ with $f = \mathtt{>}$ or $f = \mathtt{<}$, $t_0, t_1 \in T_A$. Then we define
\begin{enumerate}

\item Left subtree extraction: $\ext_0: T_A \to T_A$,
    \[
       \ext_0(t) =  t_0 \:.
    \]
\item Right subtree extraction: $\ext_1: T_A \to T_A$,
    \[
        \ext_1(t)  = t_1 \:.
    \]
\item Tree constructions: $\cons_f: T_A \times T_A \to T_A$,
    \[
        \cons_f(t_0, t_1) = t \:.
    \]
\end{enumerate}
\end{definition}
Recursion with left and right tree extraction is applied as follows:
\begin{definition}\label{def:exrecurs}
Let $I = \{0, 1 \}^*$ be the set of binary sequences, $\gamma = \gamma_1 \gamma_2 \ldots \gamma_n \in I$, for $n \in \mathbb{N}_0$.
Then the function $\ext_\gamma: T_A \to T_A$ is given as the concatenation product
\begin{eqnarray*}
  \ext_\varepsilon &=& \id \\
  \ext_{i \gamma} &=& \ext_i \circ \ext_\gamma \:,
\end{eqnarray*}
where $\id : T_A \to T_A$ denotes the identity function, $\id(t) = t$, for all $t \in T_A$. The bit strings $\gamma \in I$ are called \emph{node addresses} for minimalist trees and $I$ is the \emph{address space}.
\end{definition}

Using node addresses we fetch the function symbols of terms through another function.
\begin{definition}\label{def:label}
Let $t \in T_A$ be given as $t = f(t_0, t_1)$ with $f = \mathtt{>}$ or $f = \mathtt{<}$, $t_0, t_1 \in T_A$, and $\gamma \in I$. Then $\lab : I \times T_A \to A$ with
\begin{eqnarray*}
 \lab(\varepsilon, t) &=& f \\
 \lab(i \gamma, t) &=& \lab(\gamma, \ext_i(t)) \:.
\end{eqnarray*}
If $t$ is a constant in $T_A$, however (i.e. $t \in T_F$), then
\[
 \lab(\gamma, t) = t \:,
\]
for every $\gamma \in I$.
\end{definition}

\begin{corollary}\label{col:treecons}
As a collorary of definitions \ref{def:treefnc1} and \ref{def:label} we state
 \begin{equation}\label{eq:conslabel}
    t = \cons_{\lab(\varepsilon, t)}(\ext_0(t), \ext_1(t)) \:,
 \end{equation}
if $\rank_A(\lab(\varepsilon, t)) = 2$ for $t \in T_A$.
\end{corollary}

\begin{definition}\label{def:head}
The \emph{head} of a minimalist tree $t \in T_A$ is a unique leaf that projects over all other nodes of the tree. We find $t$'s head address by recursively following the projection labels. Therefore, $\head:T_A \to I$ is defined through
\begin{eqnarray*}
  \head(<(t_0, t_1)) &=& 0^\frown \head(t_0) \\
  \head(>(t_0, t_1)) &=& 1^\frown \head(t_1) \:,
\end{eqnarray*}
where string concatention is indicated by ``$^\frown$'', and
\[
  \head(t) =  \varepsilon \:,
\]
for $t \in T_F$
\end{definition}

\begin{definition}\label{def:feature}
The \emph{feature} of a tree $t$ is defined as the first feature of $t$'s head label. Thus $\feat: T_A \to F_F$,
\[
 \feat(t) = \first(\lab(\head(t), t))  \:,
\]
where we appropriately extended domain and codomain of $\first: A \setminus \{ \varepsilon \} \to F_F \cup \{ \varepsilon \} $, by setting $\first(<) = \first(>) = \varepsilon$.
\end{definition}

A node in a minimalist tree $t$ is known to be a \emph{maximal projection} if it is either $t$'s root, or if its sister projects over that node. We exploit this property in order to recursively determine the address of a maximal subtree for a given node address.
\begin{definition}\label{def:maxpro}

Let $t \in T_A$ and  $\gamma \in I$. Then, $\max: I \times T_A \to I$,
\[
    \max(\gamma, t) = \left\{ \begin{array}{c@{\quad:\quad}l}
                         \varepsilon & \gamma = \head(t) \\
                         i^\frown \max(\delta, \ext_i(t)) &  \gamma = i \delta \mbox{ and } \gamma \ne \head(t) \\
                         \mbox{undefined} & \mbox{otherwise}
                \end{array}
         \right.
\]
is a partial function.

\end{definition}

We also need a variant thereof with wider scope. Thus we additionally define:
\begin{definition}\label{def:maxpro2}
Let $P \subset I$ be a set of node addresses, then $\max{}^\#: \wp(I) \times T_A \to \wp(I)$,
\[
    \max{}^\#(P, t) = \bigcup_{\gamma \in P} \{ \max(\gamma, t) \} \:.
\]
If $P$ is a singleton set, $P = \{\gamma\}$, we identify the actions of $\max$ and $\max{}^\#$. Here, $\wp(I)$ denotes the power set of node addresses $I$.
\end{definition}

Moreover, we define a function that returns the leaf addresses of a tree $t$ possessing the same feature $f \in F_F$.

\begin{definition}\label{def:leaves}
Let $t \in T_A$ and $f \in F_F$. Then, $\leav : F_F \times T_A \to \wp(I)$, with
\[
  \leav(f, t) = \{ \gamma \in I | \first(\lab(\gamma, t)) = f \} \:.
\]
where $\gamma$ varies over the address space of $t$.
\end{definition}

Next, we introduce a term replacement function.
\begin{definition}\label{def:replace}
Let $t, t' \in T_A$ and $\gamma \in I$. Then $\rep : I \times T_A \times T_A \to T_A$ with
\begin{eqnarray*}
  \rep(\varepsilon, t, t') &=& t' \\
  \rep(0 \gamma, t, t') &=& \cons_{\lab(\varepsilon, t)}(\rep(\gamma, \ext_0(t), t'), \ext_1(t)) \\
  \rep(1 \gamma, t, t') &=& \cons_{\lab(\varepsilon, t)}(\ext_0(t), \rep(\gamma, \ext_1(t), t')) \:.
\end{eqnarray*}
\end{definition}
Using replace we extend the domain of the shift function (\ref{def:strfnc}) from the string algebra $T_F$ to the tree algebra $T_A$.
\begin{definition}\label{def:shift2}
Let $t \in T_A$. Then, $\shift^\#: T_A \to T_A$ with
\[
  \shift^\#(t) = \rep(\head(t), t, \shift(\lab(\head(t), t))) \:,
\]
deletes the first feature of $t$'s head.
\end{definition}

\begin{figure}[H]
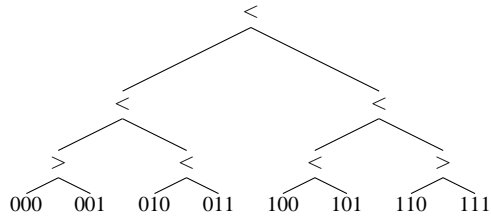

\Tree [.$<$
        [.$<$
            [.$>$
                $000$ $001$
            ].$>$
            [.$<$
                $010$ $011$
            ].$<$
        ].$<$
        [.$<$
            [.$<$
                $100$ $101$
            ].$<$
            [.$>$
                $110$ $111$
            ].$>$
        ].$<$
    ].$<$
 \caption{\label{fig:tree2} Labeled minimalist tree $t$ with leaf addresses for illustration of $\head$ and $\max$ functions: $001 = \head(t)$ and, e.g., $\max(100, t) = 1$.}
\end{figure}

The effect of the tree functions $\head$ and $\max$ are illustrated in \Fig{fig:tree2}. The head of the tree $t$ is obtained by following the projection indicators recursively through the tree: $\head(t) = 0^\frown\head(\ext_0(t)) = 00^\frown\head(\ext_0(\ext_0((t)))) = 001^\frown\head(\ext_1(\ext_0(\ext_0((t))))) = 001$; and $\max(100, t) = 1^\frown\max(00, \ext_1(t)) = 1$.

% ------------------------------------- Section -----------------------------------
\subsection{Minimalist grammars}
\label{sec:mg2}

Now we are prepared to define minimalist grammars in term algebraic terms.

\begin{definition}\label{def:mg}
A minimalist grammar (MG) is a four-tuple $G = (P, C, \Lex, \mathcal{M})$ obeying conditions (1) -- (4).
\begin{enumerate}
\item $P$ is a finite set of non-syntactic \emph{phonetic} features.
\item $C = B \cup S \cup L \cup M$ is a finite set of syntactic features, called \emph{categories}, comprising \emph{basic categories}, $B$, \emph{selectors}, $S$, \emph{licensors}, $L$, and \emph{licensees}, $M$. There is one distinguished element, $\mathtt{c} \in B$, called \emph{complementizer}. $F_F = P \cup C$ is then the feature set. To each selector $s \in S$ a basic category $b \in B$ is assigned by means of a \emph{select function}, $\sel: S \to B$, $b = \sel(s)$. Likewise, a \emph{license function}, $\lic: L \to M$ assigns to each licensor $\ell \in L$ a corresponding licensee through $m = \lic(\ell)$.
\item $\Lex \subset T_F$ is a finite set of simple terms over the term algebra $T_F$, called the \emph{lexicon}, such that each term $t \in \Lex$, is a feature string of the form
    \[
        S^* (L \cup \{ \varepsilon \}) S^* B  M^* P^* \:.
    \]
\item $\mathcal{M} = \{ \merge, \move \}$ is a collection of partial functions, $\merge: T_A \times T_A \to T_A$ and $\move: T_A \to T_A$, defined as follows: The domain of merge is given by all pairs of trees $\DomMerge = \{ (t_1, t_2) \in T_A \times T_A | \sel(\feat(t_1)) = \feat(t_2) \}$. The domain of move contains all trees $\DomMove = \{ t \in T_A | \feat(t) \in L \text{ and } \max{}^\#(\leav(\lic(\feat(t)), t), t) \\
    \text{contains exactly one element} \}$. Let $t_1, t_2 \in \DomMerge$ and $t \in \DomMove$, then
    \begin{eqnarray*}
      \merge(t_1, t_2) &=& \left\{ \begin{array}{c@{\mbox{ if }}l}
                    \cons_<(\shift^\#(t_1), \shift^\#(t_2)) & t_1 \text{ is simple} \\
                    \cons_>(\shift^\#(t_1), \shift^\#(t_2)) & t_1 \text{ is complex}
                                        \end{array}
         \right. \\
      \move(t) &=& \cons_>( \shift^\#(\ext_{\max(\leav(\lic(\feat(t)), t), t)}(t)), \\
        && \shift^\#(\rep(\max(\leav(\lic(\feat(t)), t), t), t, \varepsilon)))
      )
    \end{eqnarray*}
\end{enumerate}
\end{definition}

The constraint on the move operation, that the set of maximal subtrees with the corresponding licensee may contain exactly one element is called the \emph{shortest move condition}, motivated by linguistic considerations. Relaxing this condition yields different kinds of minimalist grammars that could account for particular locality conditions \citep{GartnerMichaelis07a}.

% ------------------------------------- Section -----------------------------------
\subsection{Processing algorithm}
\label{sec:pars}

Minimalist grammar recognition and parsing are well understood \citep{Harkema01, Mainguy10, Stabler11b}. However, for our current exposition, instead of a full-fletched minimalist parser that must be proven to be sound and complete, we discuss a simplified processor for our particular example from \Sec{sec:applMG} below, just in order to provide a proof-of-concept for our representation theory. To this end we utilize early ideas of \citet{Stabler96} as employed by \citet{Gerth06} and \citet{GerthGraben09b}. There, the structure building functions $\merge$ and $\move$ are extended to a \emph{state description}, or a \emph{stack}, regarded as a finite word of terms $w \in T_A^*$. From a graph theoretical point of view, a state description is an unconnected collection of trees, and therefore a \emph{forest}. In order to construct an algorithm that generates a successful derivation we introduce the following extensions of $\merge$ and $\move$ over forests of minimalist trees.
\begin{definition}\label{def:pars}
Let $w = (w_1, w_2, \ldots, w_m) \in T_A^*$ be state description. Then
\begin{enumerate}
\item $\merge^*: T_A^* \to T_A^*$ with $\merge^*(w) = (w_1, w_2, \ldots, \merge(w_{m-1}, w_m))$, when $(w_{m-1}, w_m) \in \DomMerge$.
\item $\move^*: T_A^* \to T_A^*$ with $\move^*(w) = (w_1, w_2, \ldots, \move(w_m))$, when $w_m \in \DomMove$.
\end{enumerate}
are partial functions acting upon state descriptions from $T_A^*$.
\end{definition}

In \Def{def:pars}, $\merge^*$ operates on the next-to-last and the last element of the processor's state description, respectively, thereby implementing a stack with the last element at the top. Using this convention, canonical subject-verb-object sentences, [S[VO]], such as the example below and also examples used by \citet{Gerth06} and \citet{GerthGraben09b}, can be processed straightforwardly, by first merging the verb V with the direct object O as its complement, and subsequently merging the result with the subject noun phrase. Thereby, the procedure avoids unnecessary garden-path interpretations. However, since minimalist languages cannot be processed with simple pushdown automata, one needs additional mechanisms such as indices and sorting in the framework of multiple context-free languages for which the crucial soundness and completeness properties of minimalist parsers have been proven \citep{Mainguy10, Stabler11b}. In our simplified approach, however, we make use of an oracle for rearranging stack content. This is implemented through suitable permutations $\pi : T_A^* \to T_A^*$, acting upon the stack according to $w' = \pi(w)$.

The processor operates in several loops: two for the domain of merge and another one for the domain of move. In the loops for the domain of merge the iteration starts with the tree on top of the stack which is checked against every other tree whether they can be merged, in which case an appropriate permutation brings both trees into the last and next-to-last position of the stack. Then $\merge^*$ is applied and this loop iteration is terminated. If the top tree cannot be merged then the algorithm decrements backwards until it reaches the first tree on the stack. In the loop for the domain of move every tree is checked for being in the domain of move, in this case the $\move^*$ operation is used after a permutation bringing that tree into the last position of the stack. The rest of the lexical entries in the state description are passed on unchanged to the next state of the algorithm.

Therefore after $\merge^*$ or $\move^*$ has been applied to the state description the algorithm completes the current state and continues with the next one resulting in a sequence of state descriptions $S_0, S_1, ...$ which describes the derivation process. The algorithm stops when no further $\merge^*$ or $\move^*$ is applicable and only one tree remains in the state description. This final state description determines the successful derivation.

% ------------------------------------- Section -----------------------------------
\subsection{Application}
\label{sec:applMG}

We illustrate the procedure from \Def{def:pars} by constructing a minimalist grammar for the following English sentence and by outlining a successful derivation of

 \ex. Douglas loved deadlines.\footnote{Douglas Adams was quoted as saying: ``I love deadlines. I like the whooshing sound they make as they fly by." in Simpson, M. J. (2003). \textit{Hitchhiker: A Biography of Douglas Adams.} Justin, Charles and Co.,  Boston (MA).}\label{ex:adams}

The minimalist lexicon is shown in \Fig{fig:mg}. The first item is a complementizer (basic category \texttt{c}) which selects tense (indicated by the feature $\tt{=t}$). The second item is a determiner phrase ``Douglas'' (basic category $\tt{d}$) requiring case (licensee $\tt{-case}$). The third item, the verb ``love" (category $\tt{v}$), selecting a determiner (feature $\tt{=d}$), is a verb (feature $\tt{v}$) and is moved into the position before ``-ed" triggered by $\tt{-i}$ resulting in the inflection of the verb (i.e., ``loved"). The next item would normally include an affix (e.g., -ven, -ing) but it is empty ($\varepsilon$) here, it selects a verb (feature $\tt{=v}$), a determiner phrase (feature $\tt{=d}$) to which it assigns case (feature $\tt{+CASE}$) and has the feature $\tt{v}$. The fifth item represents the past tense inflection ``-ed" with the category $\tt{t}$ that selects a verb ($\tt{=v}$), assigns case (licensor $\tt{+CASE}$) to a determiner and contains the licensor $\tt{+I}$ to trigger the movement of ``love". The last item in the lexicon is the object ``deadlines" (category $\tt{d}$) which requires case ($\tt{-case}$).
\begin{figure}[H]
\centering
\[
\begin{bmatrix}
  \tt{=t} \\ \mbox{c} \\
\end{bmatrix}
\begin{bmatrix}
  \tt{d} \\  \tt{-case} \\  \mbox{Douglas} \\
\end{bmatrix}
\begin{bmatrix}
  \tt{=d} \\ \tt{v} \\ \tt{-i} \\ \mbox{love} \\
\end{bmatrix}
\begin{bmatrix}
  \tt{=v} \\ \tt{+CASE} \\ \tt{=d} \\ \tt{v} \\ \mbox{$\epsilon$} \\
\end{bmatrix}
\begin{bmatrix}
  \tt{=v} \\ \tt{+I} \\ \tt{+CASE} \\ \tt{t} \\ \mbox{-ed} \\
\end{bmatrix}
\begin{bmatrix}
  \tt{d} \\ \tt{-case} \\ \mbox{deadlines} \\
\end{bmatrix}
\]
\caption{Minimalist lexicon of sentence \ref{ex:adams}.}\label{fig:mg}
\end{figure}

The algorithm takes initially as input the state description $w_1$ = (Douglas, love, -ed, deadlines) $\in T_A^*$.

\subsubsection{An example derivation of sentence \ref{ex:adams}}
\begin{figure}[H]
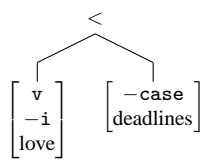

\centering
 \Tree [.$<$ [ {$\begin{bmatrix} \tt{v} \\ \tt{-i} \\ \mbox{love} \\ \end{bmatrix}$} ]
		  [ {$\begin{bmatrix} \tt{-case} \\ \mbox{deadlines} \\ \end{bmatrix}$} ]
	 ].$<$
\caption{Step 1: merge.}\label{fig:step1}
\end{figure}

Starting with the initial state description $w_1$ the words ``love" and ``deadlines" are merged (\Fig{fig:step1}) after a first permutation $\pi_1$, exchanging ``-ed'' and ``love'', by applying $\merge^*(\pi_1(w_1))$ =(Douglas, -ed, $\merge$(love, deadlines)) because ``love" ($\tt{=d}$) and ``deadlines" ($\tt{d}$) are in $\DomMerge$.

In the next step $\epsilon$ is merged to the tree.
\begin{figure}[H]
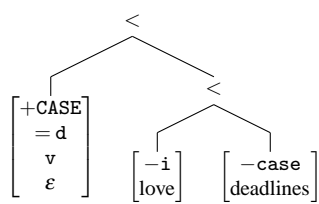

\[
  \Tree [.$<$ [ {$\begin{bmatrix} \tt{+CASE} \\ \tt{=d} \\ \tt{v} \\ \mbox{$\epsilon$} \\ \end{bmatrix}$} ]
  	[.$<$ [ {$\begin{bmatrix} \tt{-i} \\ \mbox{love} \\ \end{bmatrix}$} ]
		  [ {$\begin{bmatrix} \tt{-case} \\ \mbox{deadlines} \\ \end{bmatrix}$} ]
	 ].$<$
	].$<$
\]
\caption{Step 2: merge.}\label{fig:step2}
\end{figure}

\begin{figure}[H]
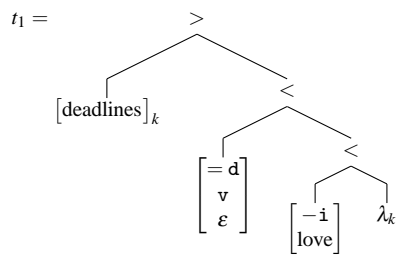

\[
t_1 =  \Tree [.$>$ [ {$\begin{bmatrix} \mbox{deadlines} \\ \end{bmatrix}_k$} ]
  	[.$<$ [ {$\begin{bmatrix} \tt{=d} \\ \tt{v} \\ \mbox{$\epsilon$} \\ \end{bmatrix}$} ]
  	[.$<$ [ {$\begin{bmatrix} \tt{-i} \\ \mbox{love} \\ \end{bmatrix}$} ]
		  [ {$\lambda_k$} ]
	 ].$<$
	].$<$
	].$>$
\]
\caption{Step 3: move.}\label{fig:step3}
\end{figure}

The resulting tree is in the domain of move triggered by the features $\tt{-case}$ and $\tt{+CASE}$, therefore ``deadlines" is moved upwards in the tree leaving behind $\lambda$, a new leaf node without label. The involved expressions are co-indexed with $k$ (\Fig{fig:step3}).

In step 4 the whole state description $w_2 = (\text{Douglas}, \text{-ed}, t_1)$ is checked for being in the domain of merge. This is the case for $(\text{Douglas}, t_1)$. Therefore, ``Douglas'' is merged to $t_{1}$.
\begin{figure}[H]
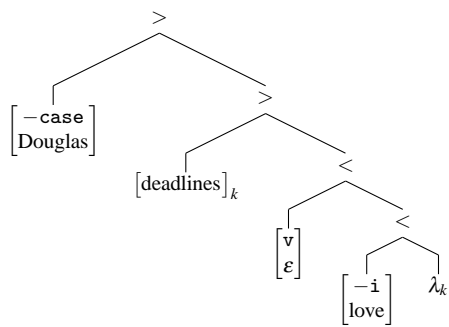

\centering
\Tree [.$>$ [ {$\begin{bmatrix} \tt{-case} \\ \mbox{Douglas} \\ \end{bmatrix}$} ]
	[.$>$ [ {$\begin{bmatrix} \mbox{deadlines} \\ \end{bmatrix}_k$} ]
  	[.$<$ [ {$\begin{bmatrix} \tt{v} \\ \mbox{$\epsilon$} \\ \end{bmatrix}$} ]
  	[.$<$ [ {$\begin{bmatrix} \tt{-i} \\ \mbox{love} \\ \end{bmatrix}$} ]
		  [ {$\lambda_k$} ]
	 ].$<$
	].$<$
	].$>$
	].$>$
\caption{Step 4: merge.}\label{fig:step4}
\end{figure}

Next, the past tense inflection ``-ed" is merged to the tree triggered by $\tt{v}$.
\begin{figure}[H]
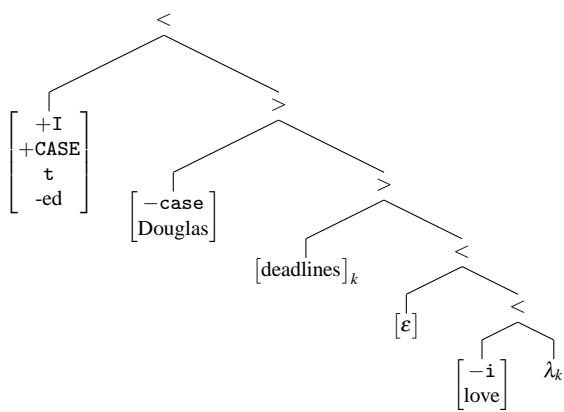

\centering
\Tree [.$<$ [ {$\begin{bmatrix} \tt{+I} \\ \tt{+CASE} \\ \tt{t} \\ \mbox{-ed} \\ \end{bmatrix}$} ]
	[.$>$ [ {$\begin{bmatrix} \tt{-case} \\ \mbox{Douglas} \\ \end{bmatrix}$} ]
	[.$>$ [ {$\begin{bmatrix} \mbox{deadlines} \\ \end{bmatrix}_k$} ]
  	[.$<$ [ {$\begin{bmatrix} \mbox{$\epsilon$} \\ \end{bmatrix}$} ]
  	[.$<$ [ {$\begin{bmatrix} \tt{-i} \\ \mbox{love} \\ \end{bmatrix}$} ]
		  [ {$\lambda_k$} ]
	 ].$<$
	].$<$
	].$>$
	].$>$
	].$<$
\caption{Step 5: merge.}\label{fig:step5}
\end{figure}

\begin{figure}[H]
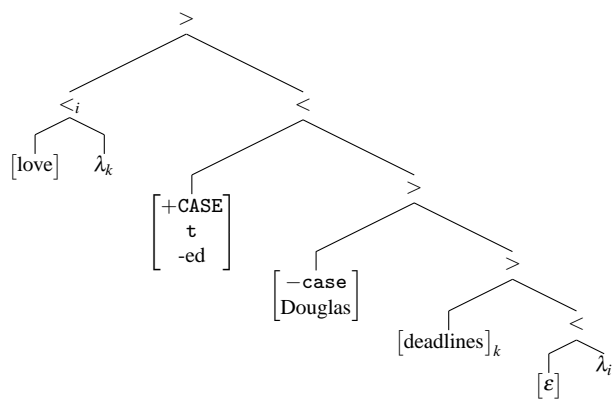

\centering
\Tree [.$>$
    [.$<_i$ [ {$\begin{bmatrix} \mbox{love} \\ \end{bmatrix}$} ]
		  [ {$\lambda_k$} ]
	 ].$<_i$
    [.$<$ [ {$\begin{bmatrix} \tt{+CASE} \\ \tt{t} \\ \mbox{-ed} \\ \end{bmatrix}$} ]
	[.$>$ [ {$\begin{bmatrix} \tt{-case} \\ \mbox{Douglas} \\ \end{bmatrix}$} ]
	[.$>$ [ {$\begin{bmatrix} \mbox{deadlines} \\ \end{bmatrix}_k$} ]
  	[.$<$ [ {$\begin{bmatrix} \mbox{$\epsilon$} \\ \end{bmatrix}$} ]
  $\lambda_i$	
	].$<$
	].$>$
	].$>$
	].$<$
].$>$
\caption{Step 6: move.}\label{fig:step6}
\end{figure}

Now, the tree is in the domain of move triggered by $\tt{-i}$ and $\tt{+I}$. Therefore, the maximal projection $\text{love} \: \lambda_k$ undergoes remnant movement to the specifier position in
\Fig{fig:step6}.

\begin{figure}[H]
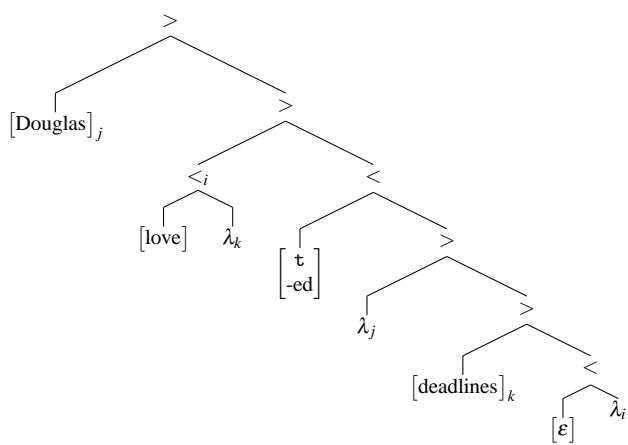

\centering
\Tree [.$>$ [ {$\begin{bmatrix} \mbox{Douglas} \\ \end{bmatrix}_j$} ]
	[.$>$
    [.$<_i$ [ {$\begin{bmatrix} \mbox{love} \\ \end{bmatrix}$} ]
		  [ {$\lambda_k$} ]
	 ].$<_i$
    [.$<$ [ {$\begin{bmatrix} \tt{t} \\ \mbox{-ed} \\ \end{bmatrix}$} ]
	[.$>$ [ {$\lambda_j$} ]
	[.$>$ [ {$\begin{bmatrix} \mbox{deadlines} \\ \end{bmatrix}_k$} ]
  	[.$<$ [ {$\begin{bmatrix} \mbox{$\epsilon$} \\ \end{bmatrix}$} ]
  $\lambda_i$	
	].$<$
	].$>$
	].$>$
	].$<$
].$>$
	].$>$
\caption{Step 7: move.}\label{fig:step7}
\end{figure}

The resulting tree is again in $\DomMove$ and ``Douglas" is moved upwards leaving a $\lambda$ behind indexed with $j$ (\Fig{fig:step7}).

\begin{figure}[H]
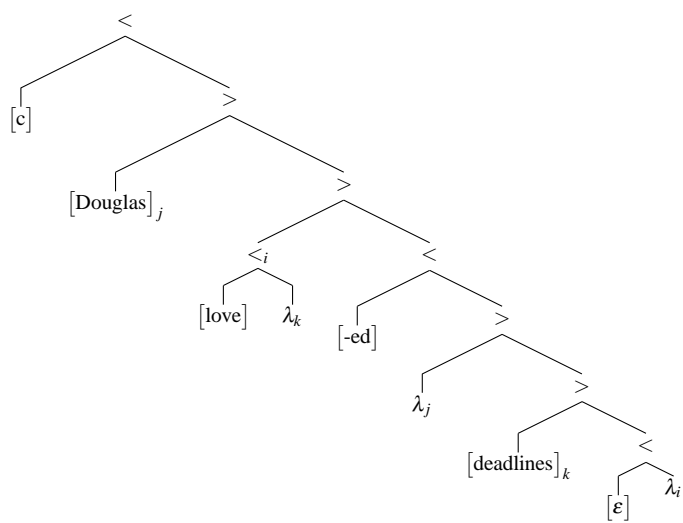

\centering
\Tree [.$<$ [ {$\begin{bmatrix} \mbox{c} \\ \end{bmatrix}$} ]
	[.$>$ [ {$\begin{bmatrix} \mbox{Douglas} \\ \end{bmatrix}_j$} ]
	[.$>$
    [.$<_i$ [ {$\begin{bmatrix} \mbox{love} \\ \end{bmatrix}$} ]
		  [ {$\lambda_k$} ]
	 ].$<_i$
    [.$<$ [ {$\begin{bmatrix} \mbox{-ed} \\ \end{bmatrix}$} ]
	[.$>$ [ {$\lambda_j$} ]
	[.$>$ [ {$\begin{bmatrix} \mbox{deadlines} \\ \end{bmatrix}_k$} ]
  	[.$<$ [ {$\begin{bmatrix} \mbox{$\epsilon$} \\ \end{bmatrix}$} ]
  $\lambda_i$	
	].$<$
	].$>$
	].$>$
	].$<$
].$>$
	].$>$
	].$<$
\caption{Step 8: merge.}\label{fig:step8}
\end{figure}

In the final step, the complementizer ``c" is merged to the tree leading to the final minimalist tree with the unchecked feature $\tt{c}$ as its head (\Fig{fig:step8}) that completes the successful derivation.

% ------------------------------------- Section -----------------------------------
\section{Integrated Symbolic/Connectionist Architectures}
\label{sec:isc}

Connectionist models of symbolic computations are an important branch in cognitive science. In order to construe compositional representations \citep{FodorPylyshyn88} one has to solve the famous \emph{binding problem} known from the neurosciences \citep{EngelEA97}: How are representations from different perceptual modalities bound together in the representation of a complex concept? The same problem appears for complex data structures such as lists or trees, e.g., in computational linguistics \citep{Hagoort05}: How is a syntactic category bound to its functional role in a phrase structure tree?

A solution for this binding problem has been provided by Smolensky's Integrated Connectionist/Symbolic Architectures (ICS) \citep{Smolensky06, SmolenskyLegendre06a, SmolenskyLegendre06b}. Here, complex symbolic data structures are decomposed into content fillers and functional roles that bind together in a geometric representation by means of tensor products. A closely related approach is Dynamic Cognitive Modeling (DCM) \citep{GrabenPotthast09a, GrabenPotthastTA}, where neural network models are explicitly constructed from geometric representations by solving inverse problems \citep{PotthastGraben09a}.

In this section, we apply the concepts of ICS/DCM to our reconstruction of minimalist grammars and processor, obtained in \Sec{sec:mg}.

% ------------------------------------- Section -----------------------------------
\subsection{Filler/role bindings}
\label{sec:frbind}

Consider a set of symbolic structures $S$ and some structure $s \in S$. A filler/role binding of $s$ is then a set of ordered pairs $\beta(s)$ of fillers bound to roles.

\begin{definition}\label{def:frbind}
Let $F$ be a finite set of \emph{simple fillers} and $R$ be a finite, countable, or even measurable set of \emph{roles}. By induction we define a family of \emph{complex fillers} as follows:
\begin{eqnarray*}
\label{eq:frbindn+1}
 F_0 &=& F \\
 F_{n + 1} &=& \wp(F_n \times R) \nonumber \:,
\end{eqnarray*}
where $n \in \mathbb{N}_0$ and $\wp(X)$ denotes the power set of some set $X$. Furthermore we define the collection
\[
 F_\infty = R \cup \left( \bigcup_{n = 0}^\infty F_n \right) \:.
\]
The \emph{filler/role binding} for $S$ is a mapping $\beta : S \to F_\infty$.
\end{definition}

In the simplest case, simple fillers are bound to roles. Thus, a filler/role binding $\beta(s) = \{(f, r) | f \in F, r \in R \} \in \wp(F \times R) = F_1$. Such a decomposition could act as a complex filler $f'$ for another filler/role binding where $f' = \beta(s)$ is bound to a role $r$, resulting in $\beta(s') = \{ (f', r) | f' \in F_1, r \in R \} \in \wp(F_1 \times R) = F_2$. By means of recursion any finite structure of arbitrary complexity yields its filler/role binding as an element of $F_\infty$ \citep{GrabenEA08b}.

Next we construct filler/role bindings for minimalist trees, $S = T_A$, in a hierarchical manner. To this aim we start with feature strings.

% ------------------------------------- Section -----------------------------------
\subsubsection{Feature strings}
\label{sec:featstrics}

Let $S = T_F$ be the string term algebra over signature  $F_F \cup \{ \varepsilon \}$ from \Sec{sec:featstr}. A string $s = (f_1 \circ f_2 \circ \ldots \circ f_p)(\varepsilon) \in T_F$  assumes a straightforward filler/role binding by interpreting $F_F$ as the filler set. Then each string position $i$ is identified with one role, $s_i \in R_F$, such that $R_F = \{ s_i | i \in \mathbb{N} \}$ is an infinite but countable set of roles. However, since every string $s \in T_F$ is of finite length $p$, only roles from $R_p = \{ s_i | 1 \le i \le p \}$ are required.

\begin{definition}\label{def:strfr}
An order-reverting filler/role binding $\beta_F: T_F \to \wp(F_F \times R_F) $ for feature string $s = f(r) \in T_F$ of length $p > 0$ is given as a mapping
\begin{eqnarray*}
\beta_F(\varepsilon) &=& \emptyset \\
\beta_F(f(r)) &=& \{ (f, s_p) \} \cup \beta_F(r) \:.
\end{eqnarray*}
\end{definition}
As an example consider the term $(f_1 \circ f_2)(\varepsilon) \in T_F$. Its filler/role binding is then
\[
 \beta_F(f_1(f_2(\varepsilon))) = \{ (f_1, s_2) \} \cup  \beta_F(f_2(\varepsilon)) =
 \{ (f_1, s_2) \} \cup  \{ (f_2, s_1) \} \cup \beta_F(\varepsilon) = \{ (f_1, s_2), (f_2, s_1) \} \:.
\]

% ------------------------------------- Section -----------------------------------
\subsubsection{Labeled trees}
\label{sec:labtreeics}

\begin{figure}[H]
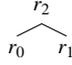

\Tree [.$r_2$ $r_0$ $r_1$ ]
 \caption{\label{fig:treeroles} Elementary roles of a labeled binary tree.}
\end{figure}

The filler/role binding for labeled binary trees has been discussed by \citet{GrabenGerthVasishth08, GrabenEA08b}, and \citet{GrabenPotthast09a}. For tree term algebras $T_A$ from \Sec{sec:labtree}, we identify the signature $A = T_F \cup \{ \mathtt{<},  \mathtt{>} \}$ with the set of simple fillers and introduce roles $R_A = \{r_0, r_1, r_2\}$, with ``mother'' ($r_2$), ``left daughter'' ($r_0$) and ``right daughter'' ($r_1$) of an elementary tree as indicated in \Fig{fig:treeroles}, where the indices have been chosen in accordance with the extraction functions $\ext_0$ and $\ext_1$ from \Def{def:treefnc1}, such that $\ext_0(t)$ is bound to role $r_0$ and $\ext_1(t)$ is bound to role $r_1$ for a term $t \in T_A$. In accordance to \Def{def:frbind}, we call the set of complex fillers $A_\infty$. Additionally, we unify the sets of simple fillers and roles through
\begin{eqnarray}
 \label{eq:treefilrole1}  F &=&  F_F \cup \{ \mathtt{<},  \mathtt{>} \} \\
 \label{eq:treefilrole2}   R &=& R_F \cup R_A \:.
\end{eqnarray}

\begin{definition}\label{def:treefr}
A filler/role binding $\beta_A: T_A \to A_\infty$ for tree terms is given as a mapping
\[
\beta_A(t) = \left\{ \begin{array}{c@{\mbox{ if }}l}
                         \{ (f, r_2), (\beta_A(t_0), r_0), (\beta_A(t_1), r_1) \} & t = f(t_0, t_1) \in T_A \\
                        \beta_F(t) & t \in T_F \:.
                             \end{array}
         \right.
\]
\end{definition}

\begin{figure}[H]
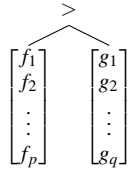

\Tree [.$>$
    {$\begin{bmatrix} f_1 \\ f_2 \\ \vdots \\ f_p \end{bmatrix}$}
    {$\begin{bmatrix} g_1 \\ g_2 \\ \vdots \\ g_q \end{bmatrix}$} ]
 \caption{\label{fig:simpexpr} Minimalist tree term $t = \mathtt{>}(f, g) \in T_A$ with feature $g_1$.}
\end{figure}

Consider the minimalist tree $t = \mathtt{>}(f, g) \in T_A$ in \Fig{fig:simpexpr} where the root is labeled with the projection indicator pointing to the head at the right daughter and feature string terms $f = (f_1 \circ f_2 \circ \ldots \circ f_p)(\varepsilon) \in T_F$, $p \in \mathbb{N}$, $g = (g_1 \circ g_2 \circ \ldots \circ g_q)(\varepsilon) \in T_F$, $q \in \mathbb{N}$, are presented as column arrays.

The filler/role binding of $t$ is obtained as
\begin{multline*}
\beta_A(t) = \beta_A(\mathtt{>}(f, g)) =
 \{ (\mathtt{>}, r_2), (\beta_A(f), r_0), (\beta_A(g), r_1) \} =
 \{ (\mathtt{>}, r_2), (\beta_F(f), r_0), (\beta_F(g), r_1) \} = \\
=  \{ (\mathtt{>}, r_2),
    (\{ (f_1, s_p), (f_2, s_{p-1}), \ldots, (f_p, s_1) \}, r_0),
    (\{ (g_1, s_q), (g_2, s_{q-1}), \ldots, (g_q, s_1) \}, r_1)
 \} \:.
\end{multline*}

\begin{figure}[H]
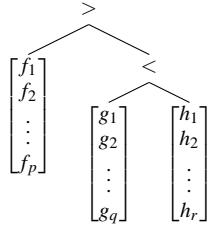

\Tree
 [.$>$
  	{$\begin{bmatrix} f_1 \\ f_2 \\ \vdots \\ f_p \end{bmatrix}$}
	[.$<$
        {$\begin{bmatrix} g_1 \\ g_2 \\ \vdots \\ g_q \end{bmatrix}$}
        {$\begin{bmatrix} h_1 \\ h_2 \\ \vdots \\ h_r \end{bmatrix}$}
  	].$<$
  ].$>$
 \caption{\label{fig:compexpr} Complex minimalist tree $s = \mathtt{>}(f, \mathtt{<}(g, h)) \in T_A$ with feature $g_1$.}
\end{figure}

A more complex expression $s = \mathtt{>}(f, \mathtt{<}(g, h)) \in T_A$ is shown in \Fig{fig:compexpr}.

The filler/role binding for the term $s$ in \Fig{fig:compexpr} is recursively constructed through
\begin{multline*}
\beta_A(s) = \beta_A(\mathtt{>}(f, \mathtt{<}(g, h))) =
 \{ (\mathtt{>}, r_2), (\beta_A(f), r_0), (\beta_A(\mathtt{<}(g, h)), r_1) \} = \\
 = \{ (\mathtt{>}, r_2), (\beta_F(f), r_0),
 ( \{ (\mathtt{<}, r_2), (\beta_A(g), r_0), (\beta_A(h), r_1) \}, r_1) \} = \\
  = \{ (\mathtt{>}, r_2), (\beta_F(f), r_0),
 ( \{ (\mathtt{<}, r_2), (\beta_F(g), r_0), (\beta_F(h), r_1) \}, r_1) \} = \\
= \{
        (\mathtt{>}, r_2),
        (\{ (f_1, s_p), (f_2, s_{p-1}), \ldots, (f_p, s_1) \}, r_0), \\
        (\{
        (\mathtt{<}, r_2),
        (\{ (g_1, s_q), (g_2, s_{q-1}), \ldots, (g_q, s_1) \}, r_0), \\
        (\{ (h_1, s_r), (h_2, s_{r-1}), \ldots, (h_r, s_1) \}, r_1)
  \}, r_1)
  \}  \:.
\end{multline*}

% ------------------------------------- Section -----------------------------------
\subsection{Tensor product representations}
\label{sec:tpr}

\begin{definition}\label{def:tpr}
Let $\mathcal{F}$ be a vector space over the real or complex numbers, and $\beta: S \to F_\infty$ a filler/role binding for a set of symbolic structures $S$ for sets of fillers $F$ and roles $R$. A mapping $\psi: F_\infty \to \mathcal{F}$ is called \emph{tensor product representation} of $S$ if it obeys (1) -- (3).
\begin{enumerate}
\item $\psi(F_n)$ is a subspace of $\mathcal{F}$, for all $n \in \mathbb{N}_0$,
     in particular for $F_0 = F$ is $\psi(F) = \mathcal{V}_F$ a subspace of $\mathcal{F}$,
\item $\psi(R) = \mathcal{V}_R$ is a subspace of $\mathcal{F}$,
\item $\psi(\{(f, r)\} ) = \psi(f) \otimes \psi(r)$, for filler $f \in F_n$ and role $r \in R$ ($n \in \mathbb{N}_0$).
\item $\psi(A \cup B) = \psi(A) \oplus \psi(B)$, for subsets $A, B \subset F_\infty$.
\end{enumerate}
\end{definition}

\begin{lemma}\label{lem:fock}
$\mathcal{F}$ is the \emph{Fock space}
\begin{equation}\label{eq:fockspace}
    \mathcal{F} =
    \left( \bigoplus_{n = 0}^\infty \mathcal{V}_F \otimes \mathcal{V}_R^{\otimes^n} \right) \oplus \mathcal{V}_R \:,
\end{equation}
known from quantum field theory \citep{Haag92, SmolenskyLegendre06a}.
\end{lemma}

\noindent\emph{Proof} (by induction over $n \in \mathbb{N}_0$). Let $n = 0$. Then $\mathcal{V}_F \otimes \mathcal{V}_R^{\otimes^0} = \mathcal{V}_F$ is a subspace of $\mathcal{F}$. Moreover $\mathcal{V}_R $ is a subspace of $\mathcal{F}$. Let $f \in F_n$, $r \in R$, such that $\psi(f) \in \psi(F_n)$ and $\{(f, r)\} \in F_{n+1}$ be a filler/role binding. Then $\psi(\{(f, r)\}) = \psi(f) \otimes \psi(r) \in \psi(F_{n+1})$. The direct sum of those subspaces is the Fock space $\mathcal{F}$.

By concatenating the maps $\beta, \psi$, we extend the tensor product representation $\Psi: S \to \mathcal{F}$ of the symbolic structures:
\begin{equation}\label{eq:Psi}
 \Psi(s) = \psi(\beta(s)) \:, \qquad s \in S \:.
\end{equation}

\begin{definition}\label{def:unbind}
Let $\beta, \psi$ be a filler/role binding and a tensor product representation for a structure set $S$ in Fock space $\mathcal{F}$ over fillers $F$ and roles $R$. A linear function $\upsilon_r: \mathcal{F} \to \mathcal{F}$ is called unbinding for role $r$ if
\[
 \upsilon_r(\vec{u})= \left\{ \begin{array}{c@{\quad:\quad}l}
                        \psi(f) & \vec{u} = \psi(\{(f, r)\}) \\
                        0 & \text{otherwise} \:.
                        \end{array}
         \right.
\]
\end{definition}
Unbinding functions can be established in several ways, e.g. by means of adjoint vectors or through self-addressing \citep{SmolenskyLegendre06a, Smolensky06}. Self-addressing requires that the Fock space $\mathcal{F}$ is equipped with a scalar product, turning it into a Hilbert space. However, in this paper, we use adjoint vectors, i.e. linear forms into its number field, from the dual space $\mathcal{V}_R^*$ of the respective role representation space. This requires that all filler and role vectors are linearly independent, implying \emph{faithful} tensor product representations \citep{SmolenskyLegendre06a, Smolensky06}.

Next, we define the realization of a symbolic computation.

\begin{definition}\label{def:comp}
Let $P, Q: S \to S$ be partial functions on the symbolic structures $s \in S$, such that $\mathrm{Cod}_P \subseteq \mathrm{Dom}_Q$. Two piecewise linear functions $\mathbf{P}, \mathbf{Q}: \mathcal{F} \to \mathcal{F}$ are called \emph{realizations} of the symbolic computations $P, Q$ in Fock space $\mathcal{F}$, if there is a tensor product representation $\Psi: S \to \mathcal{F}$ such that
\begin{eqnarray*}
    (\mathbf{P} \circ \Psi)(s) &=& (\Psi \circ P)(s) \\
    (\mathbf{Q} \circ \Psi)(t) &=& (\Psi \circ Q)(t)
\end{eqnarray*}
for all $s \in \mathrm{Dom}_P$, $t \in \mathrm{Dom}_Q$.
\end{definition}
Then, the realizations constitute a semigroup homomorphism and hence a semigroup representation in the sense of algebraic representation theory \citep{vdWaerden03b, GrabenPotthast09a}, because
\[
    (\mathbf{Q} \circ \mathbf{P} \circ \Psi)(s) = (\Psi \circ Q \circ P)(s) \:,
\]
for all $s \in \mathrm{Dom}_P$.

% ------------------------------------- Section -----------------------------------
\subsubsection{Feature strings}
\label{sec:featstr2}

In \Sec{sec:featstrics} we created filler/role bindings for the term algebra $T_F$ of minimalist feature strings as $\beta_F(s) = \beta((f_1 \circ f_2 \circ \ldots \circ f_p)(\varepsilon)) = \{ (f_1, s_p), (f_2, s_{p-1}), \ldots, (f_p, s_1) \}$ in reversed order by regarding the features as fillers $F_F$ and the string positions $R_F = \{ s_i | i \in \mathbb{N} \}$ as roles. Mapping all fillers $f_i \in F_F$ onto filler vectors $\vec{f}_i = \psi(f_i) \in \mathcal{V}_F$ from a vector space $\mathcal{V}_F$, and similarly all roles $s_i \in R_p$ for a string of length $p$ onto role vectors $\vec{s}_i = \psi(s_{p - i + 1}) \in \mathcal{V}_R$ from a vector space $\mathcal{V}_R$ yields a tensor product representation of feature strings in preserved order through
\begin{equation}\label{eq:tprstring}
    \Psi(s) = \psi(\beta_F(s)) = \sum_{i = 1}^p \vec{f}_i \otimes \vec{s}_i \:.
\end{equation}
However, for the sake of convenience, we extend the representation space to an embedding space spanned by the role vectors that are required for representing the longest feature strings. Let therefore $n \in \mathbb{N}$ be the maximal length of a feature string occuring in the minimalist lexicon, we bind the null vector $\vec{0}$ to all role vectors $\vec{s}_k$ for $k > p$ for a given string of length $p < n$. Then, all strings possess a unique representation \begin{equation}\label{eq:tprstring2}
    \Psi(s) = \psi(\beta_F(s)) = \sum_{i = 1}^n \vec{f}_i \otimes \vec{s}_i \:.
\end{equation}
We denote the embedding space for feature strings $\mathcal{S}$.

For this representation we have to find realizations of the string functions from \Def{def:strfnc}. To this end we need some preparatory concepts. Let $\vec{u} = \Psi(s) = \psi(\beta_F(s))$ be a tensor product representation for feature strings $s = (f_1 \circ f_2 \circ \ldots \circ f_p)(\varepsilon) \in T_F$, $f_i \in F$. For the role vectors $\vec{s}_i \in \mathcal{V}_R$ we define their adjoints $\vec{s}^+_i \in \mathcal{V}^*_R$ in the dual space $\mathcal{V}^*_R$ of $\mathcal{V}_R$, such that
\begin{equation}\label{eq:dual}
    \vec{s}^+_i (\vec{s}_k) = \delta_{ik} \:,
\end{equation}
with the Kronecker symbol $\delta_{ik} = 0(1)$ for $i \ne k$ ($i=k$), i.e. the adjoint vectors $\vec{s}^+_i$, acting as linear forms, and the duals $\vec{s}_i$ form a biorthogonal basis.

\begin{lemma}\label{lem:unbind}
$\upsilon_k(\vec{u}) = (\id \otimes \vec{s}^+_k)(\vec{u})$ is an unbinding function for role $s_k$.
\end{lemma}

\noindent\emph{Proof}.
Let $\vec{u} = \vec{f}_k \otimes \vec{s}_k$. Then
\begin{multline*}
 \upsilon_k(\vec{u}) = (\id \otimes \vec{s}^+_k)(\vec{u}) =
 (\id \otimes \vec{s}^+_k)\left( \vec{f}_k \otimes \vec{s}_k \right) = \vec{f}_k \vec{s}^+_k(\vec{s}_k) = \vec{f}_k \delta_{kk} = \vec{f}_k = \psi(f_k) \:.
\end{multline*}
Here, $\id : \mathcal{V}_F \to \mathcal{V}_F$ denotes the identity map at $\mathcal{V}_F$: $\id(\vec{f}) = \vec{f}$.

Since $\upsilon_k$ is also linear, we additionally obtain the following result: Let $\vec{u} = \sum_{i = 1}^n \vec{f}_i \otimes \vec{s}_i$. Then
\begin{multline*}
 \upsilon_k(\vec{u}) = (\id \otimes \vec{s}^+_k)(\vec{u}) =
 (\id \otimes \vec{s}^+_k)\left( \sum_{i = 1}^n \vec{f}_i \otimes \vec{s}_i \right) = \\
    = \sum_{i = 1}^n (\id \otimes \vec{s}^+_k)(\vec{f}_i \otimes \vec{s}_i) =
    \sum_{i = 1}^n \vec{f}_i \vec{s}^+_k(\vec{s}_i) =
        \sum_{i = 1}^n \vec{f}_i \delta_{ki} = \vec{f}_k = \psi(f_k) \:.
\end{multline*}

\begin{definition}\label{def:tprstrfnc}
Let $\Psi$ be a tensor product representation of terms of feature strings $T_F$ in vector space $\mathcal{S}$, and $\vec{u} = \Psi(s) = \sum_{i = 1}^n \vec{f}_i \otimes \vec{s}_i$ for $f \in F^*$.
\begin{enumerate}
\item The \emph{first feature} of $\vec{u}$ is obtained by an unbinding function $\bffirst: \mathcal{S} \to \mathcal{S}$ with
    \begin{equation}\label{eq:bffirst}
        \bffirst(\vec{u}) = (\id \otimes \vec{s}^+_1)(\vec{u})
    \end{equation}
\item A function $\bfshift: \mathcal{S} \to \mathcal{S}$ is obtained by
 \begin{equation}\label{eq:bfshift}
     \bfshift(\vec{u}) = \sum_{i = 1}^{n-1} ((\id \otimes \vec{s}^+_{i+1})(\vec{u})) \otimes \vec{s}_i + \vec{0} \otimes \vec{s}_n
 \end{equation}
\end{enumerate}
\end{definition}

\begin{lemma}\label{lem:tprstrfnc}
$\bffirst$ and $\bfshift$ are realizations of the corresponding string functions $\first$ and $\shift$ from \Def{def:strfnc}.
\end{lemma}

\noindent\emph{Proof}. Let $s = f(r) \in T_F$. Then
\[
\bffirst(\Psi(s)) = \vec{f} = \psi(f) = \Psi(\first(s)) \:.
\]
For $\bfshift(\vec{u})$ we compute
\begin{multline*}
 \bfshift(\Psi(s)) = \sum_{i = 1}^{n-1} \left((\id \otimes \vec{s}^+_{i+1}) \left( \sum_{k = 1}^n \vec{f}_k \otimes \vec{s}_k \right) \right) \otimes \vec{s}_i + \vec{0} \otimes \vec{s}_n = \\
 = \sum_{i = 1}^{n-1} \left( \sum_{k = 1}^n \vec{f}_k \vec{s}^+_{i+1} (\vec{s}_k) \right) \otimes \vec{s}_i + \vec{0} \otimes \vec{s}_n = \\
    \sum_{i = 1}^{n-1} \sum_{k = 1}^n \delta_{i+1,k} \vec{f}_k \otimes \vec{s}_i + \vec{0} \otimes \vec{s}_n =
    \sum_{i = 1}^{n-1} \vec{f}_{i+1} \otimes \vec{s}_i  + \vec{0} \otimes \vec{s}_n= \Psi(\shift(s)) \:.
\end{multline*}

% ------------------------------------- Section -----------------------------------
\subsubsection{Labeled trees}
\label{sec:labtree2}

A tensor product representation of a labeled binary tree is obtained from the respective filler/role binding in \Def{def:treefr}.

\begin{definition}\label{def:tprtree}
Let $t = f(t_0, t_1) \in T_A$ be a tree term with $f = \mathtt{>}$ or $f = \mathtt{<}$, $t_0, t_1 \in T_A$. Then
\[
\Psi(t) = \psi(\beta_A(t)) = \psi( \{ (f, r_2), (\beta_A(t_0), r_0), (\beta_A(t_1), r_1) \}) =
\vec{f} \otimes \vec{r}_2 \oplus \Psi(t_0) \otimes \vec{r}_0 \oplus \Psi(t_1) \otimes \vec{r}_1 \:,
\]
\end{definition}
where the projection indicators, $<$ and $>$ are mapped onto corresponding filler vectors $\vec{f}_< = \psi(<)$, $\vec{f}_> = \psi(>)$, and the three tree roles ``mother'', ``left daughter'', and ``right daughter'' are represented by three role vectors $\vec{r}_0 = \psi(r_0), \vec{r}_1 = \psi(r_1), \vec{r}_2 = \psi(r_2) \in \mathcal{V}_R$. Moreover, we also consider their adjoints $\vec{r}_0^+, \vec{r}_1^+, \vec{r}_2^+ \in \mathcal{V}_R^*$ from the dual space $\mathcal{V}_R^*$ for the required unbinding operations.

Using \Def{def:tprtree} together with the unified sets of fillers and roles from Eqs. \pref{eq:treefilrole1}, \pref{eq:treefilrole2} we can compute tensor product representations of minimalist trees, as those from the examples of \Sec{sec:labtree}. The tensor product representation of the tree term $t = \mathtt{>}(f, g) \in T_A$ in \Fig{fig:simpexpr} is given as

\begin{multline*}
 \Psi(t) = \psi(\beta_A(\mathtt{>}(f, g))) = \psi( \{ (\mathtt{>}, r_2), (\beta_A(f), r_0), (\beta_A(g), r_1) \}) = \\
 = \psi( \{ (\mathtt{>}, r_2), (\beta_F(f), r_0), (\beta_F(g), r_1) \}) =
\vec{f}_> \otimes \vec{r}_2 \oplus \psi(\{ (f_1, s_p), (f_2, s_{p-1}), \ldots, (f_p, s_1) \}) \otimes \vec{r}_0 \oplus \\
\oplus \psi(\{ (g_1, s_q), (g_2, s_{q-1}), \ldots, (g_q, s_1) \}) \otimes \vec{r}_1 = \\
= \vec{f}_> \otimes \vec{r}_2 \oplus
(\vec{f}_1 \otimes \vec{s}_1 \oplus
    \vec{f}_2 \otimes \vec{s}_2 \oplus
    \cdots \oplus \vec{f}_p \otimes \vec{s}_p) \otimes \vec{r}_0 \oplus
 (\vec{g}_1 \otimes \vec{s}_1 \oplus
    \vec{g}_2 \otimes \vec{s}_2 \oplus
    \cdots \oplus \vec{g}_q \otimes \vec{s}_q) \otimes \vec{r}_1
\end{multline*}
which can be simplified using tensor algebra to
\begin{equation}\label{eq:tprtree}
 \Psi(t) = \vec{f}_> \otimes \vec{r}_2 \oplus
\vec{f}_1 \otimes \vec{s}_1 \otimes \vec{r}_0 \oplus
    \vec{f}_2 \otimes \vec{s}_2 \otimes \vec{r}_0 \oplus
    \cdots \oplus \vec{f}_p \otimes \vec{s}_p \otimes \vec{r}_0 \oplus
 \vec{g}_1 \otimes \vec{s}_1 \otimes \vec{r}_1 \oplus
    \vec{g}_2 \otimes \vec{s}_2 \otimes \vec{r}_1 \oplus
    \cdots \oplus \vec{g}_q \otimes \vec{s}_q \otimes \vec{r}_1 \:.
\end{equation}

Correspondingly, we obtain for the tree term $s = \mathtt{>}(f, \mathtt{<}(g, h)) \in T_A$ depicted in \Fig{fig:compexpr} the tensor product representation
\begin{multline*}
 \Psi(s) =
 \vec{f}_> \otimes \vec{r}_2  \oplus
 \vec{f}_1 \otimes \vec{s}_1 \otimes \vec{r}_0  \oplus
    \vec{f}_2 \otimes \vec{s}_2 \otimes \vec{r}_0  \oplus
    \cdots \oplus \vec{f}_p \otimes \vec{s}_p \otimes \vec{r}_0  \oplus \\
 \oplus \vec{f}_< \otimes \vec{r}_2 \otimes \vec{r}_1  \oplus
 \vec{g}_1 \otimes \vec{s}_1 \otimes \vec{r}_0 \otimes \vec{r}_1 \oplus
    \vec{g}_2 \otimes \vec{s}_2 \otimes \vec{r}_0 \otimes \vec{r}_1 \oplus
    \cdots \oplus \vec{g}_q \otimes \vec{s}_q \otimes \vec{r}_0 \otimes \vec{r}_1 \oplus \\
    \oplus \vec{h}_1 \otimes \vec{s}_1 \otimes \vec{r}_1 \otimes \vec{r}_1 \oplus
    \vec{h}_2 \otimes \vec{s}_2 \otimes \vec{r}_1 \otimes \vec{r}_1 \oplus
    \cdots \oplus \vec{h}_r \otimes \vec{s}_r \otimes \vec{r}_1 \otimes \vec{r}_1 \:.
\end{multline*}
Interestingly, leaf addresses $\gamma = \gamma_1 \gamma_2 \dots \gamma_p \in I$, $p \in \mathbb{N}$, correspond to role multi-indices by means of the following convention
\begin{equation}\label{eq:roleaddr}
  \vec{r}_\gamma =
  \vec{r}_{\gamma_1} \otimes \vec{r}_{\gamma_2} \otimes \cdots \otimes \vec{r}_{\gamma_p} \:.
\end{equation}
Using role multi-indices $\vec{r}_\gamma$, we can introduce further generalized unbinding functions below.

Now we are prepared to define the Fock space realizations of the tree functions from \Sec{sec:mg}. Before we start with the counterparts from \Def{def:treefnc1}, we notice an interesting observation.

\begin{lemma}\label{lem:tprs}
Let $T_A$ be the term algebra of minimalist trees and $\Psi$ its tensor product representation in Fock space $\mathcal{F}$, as above. For $\vec{u} \in \mathcal{F}$ the function $\bffirst$ distinguishes between simple and complex trees.
\begin{enumerate}
\item $\vec{u} = \Psi(t)$ with $t \in T_F$ is a simple tree (i.e. a feature string). Then
\[
 \bffirst(\vec{u}) \ne 0 \:.
\]
\item $\vec{u} = \Psi(t)$ with $t = f(t_0, t_1) \in T_A$ is a complex tree. Then
\[
 \bffirst(\vec{u}) = 0 \:.
\]
\end{enumerate}
\end{lemma}

\noindent\emph{Proof}. Consider the first case: $\vec{u} = \Psi(t)$ for simple $t$, hence $\vec{u} = \sum_{i = 1}^n \vec{f}_i \otimes \vec{s}_i$. Then $\bffirst(\vec{u}) = \vec{f}_1 \ne 0$. For the second case, we have $\vec{u} = \Psi(t) = \vec{f} \otimes \vec{r}_2 \oplus \Psi(t_0) \otimes \vec{r}_0 \oplus \Psi(t_1) \otimes \vec{r}_1$. Therefore
\begin{multline*}
\bffirst(\vec{u}) = (\id \otimes \vec{s}^+_1)(\vec{u}) =
(\id \otimes \vec{s}^+_1)(\vec{f} \otimes \vec{r}_2 \oplus \Psi(t_0) \otimes \vec{r}_0 \oplus \Psi(t_1) \otimes \vec{r}_1) = \\
\vec{f} \, \vec{s}^+_1(\vec{r}_2) \oplus \Psi(t_0) \, \vec{s}^+_1(\vec{r}_0) \oplus \Psi(t_1) \, \vec{s}^+_1(\vec{r}_1) = 0 \:,
\end{multline*}
where $\id$ denotes the Fock space identity applied to the respective subspaces.\footnote{
Note that the Fock space identity $\id$ can be expressed as a direct sum of the respective subspace identities $\id = \sum_\kappa \id_\kappa$. Applying that to an arbitrary Fock space vector $\vec{u} = \sum_\lambda \vec{u}_\lambda$ yields $\id(\vec{u}) = \sum_\kappa \sum_\lambda \id_\kappa (\vec{u}_\lambda) = \vec{u}$, such that $\id_\kappa (\vec{u}_\lambda) = 0$ for $\kappa \ne \lambda$. Hence, the identities for different subspaces behave like orthogonal projectors, annihilating vectors from their orthocomplements. This observation applies also below.
}

\begin{definition}\label{def:tprtreefnc1}
Let $T_A$ be the term algebra of minimalist trees and $\Psi$ its tensor product representation in Fock space $\mathcal{F}$, as above. Moreover, let $\vec{u} \in \mathcal{F}$ with $\bffirst(\vec{u}) = 0$, i.e. the tensor product representation of a complex tree. Finally, let $\vec{u}_0, \vec{u}_1 \in \mathcal{F}$.
\begin{enumerate}
\item Left subtree extraction: $\bfext_0: \mathcal{F} \to \mathcal{F}$,
    \[
        \bfext_0(\vec{u}) = (\id \otimes \vec{r}^+_0)(\vec{u}) \:.
    \]
\item Right subtree extraction: $\bfext_1: \mathcal{F} \to \mathcal{F}$,
    \[
        \bfext_1(\vec{u}) = (\id \otimes \vec{r}^+_1)(\vec{u}) \:.
    \]
\item Tree constructions: $\bfcons_f: \mathcal{F} \times \mathcal{F} \to \mathcal{F}$,
    \[
        \bfcons_f(\vec{u}_0, \vec{u}_1) = \vec{u}_0 \otimes \vec{r}_0 \oplus \vec{u}_1 \otimes \vec{r}_1 \oplus \vec{f}_f \otimes \vec{r}_2 \:.
    \]
\end{enumerate}
\end{definition}

\begin{lemma}\label{lem:tprtreefnc1}
$\bfext_0$, $\bfext_1$ and $\bfcons_f$ are realizations of the corresponding string functions $\ext_0$, $\ext_1$ and $\cons_f$ from \Def{def:treefnc1}.
\end{lemma}

\noindent\emph{Proof}. Suppose $t = f(t_0, t_1),  t_0, t_1 \in T_A$. Then
\begin{multline*}
 \bfext_0(\Psi(t)) = (\id \otimes \vec{r}^+_0)(\Psi(f(t_0, t_1))) =
 (\id \otimes \vec{r}^+_0)(\vec{f} \otimes \vec{r}_2 \oplus \psi(\beta_A(t_0)) \otimes \vec{r}_0 \oplus \psi(\beta_A(t_1)) \otimes \vec{r}_1) = \\
 \vec{f} \, \vec{r}^+_0(\vec{r}_2) \oplus \psi(\beta_A(t_0)) \, \vec{r}^+_0(\vec{r}_0) \oplus \psi(\beta_A(t_1)) \, \vec{r}^+_0(\vec{r}_1) = \psi(\beta_A(t_0)) =
 \Psi(t_0) = \Psi(\ext_0(t)) \:.
\end{multline*}
The proof for $\bfext_1$ works similarly. Furthermore,
\begin{multline*}
\bfcons_f(\Psi(t_0), \Psi(t_1)) = \Psi(t_0) \otimes \vec{r}_0 \oplus \Psi(t_1) \otimes \vec{r}_1 \oplus \vec{f}_f \otimes \vec{r}_2 =
    \Psi(\cons_f(t_0, t_1)) \:.
\end{multline*}

Next, we extend those functions to node addresses as in \Def{def:exrecurs}.

\begin{definition}\label{def:tprexrecurs}
Let $I = \{0, 1 \}^*$ be the set of binary sequences, $\gamma = \gamma_1 \gamma_2 \ldots \gamma_n \in I$, for $n \in \mathbb{N}_0$. Then the function $\bfext_\gamma: \mathcal{F} \to \mathcal{F} $ is given as the concatenation product
\begin{eqnarray*}
  \bfext_\varepsilon &=& \id \\
  \bfext_{i \gamma} &=& \bfext_i \circ \bfext_\gamma \:.
\end{eqnarray*}
\end{definition}
Then, we get the following corollary from Lemma \ref{lem:tprtreefnc1}.
\begin{corollary}\label{col:tprexrecurs}
Functions $\bfext_\gamma$ are realizations of the corresponding string functions from \Def{def:exrecurs}.
\end{corollary}

Next, we realize the symbolic label function in Fock space.

\begin{definition}\label{def:tprlabel}
Let $T_A$ be the term algebra of minimalist trees and $\Psi$ its tensor product representation in Fock space $\mathcal{F}$, as above. Moreover, let $\vec{u} \in \mathcal{F}$ and $\gamma \in I$. Then $\bflab : I \times \mathcal{F}  \to \mathcal{F}$  with
\begin{eqnarray*}
 \bflab(\varepsilon, \vec{u}) &=&  (\id \otimes \vec{r}^+_2)(\vec{u}) \\
 \bflab(i \gamma, \vec{u}) &=& \bflab(\gamma, \bfext_i(\vec{u})) \:,
\end{eqnarray*}
when $\bffirst(\vec{u}) = 0$. If $\bffirst(\vec{u}) \ne 0$, then
\[
 \bflab(\gamma, \vec{u}) = \vec{u} \:.
\]
\end{definition}

\begin{lemma}\label{lem:tprlabel}
$\bflab$ is a Fock space realization of the term algebra function $\lab$ from \Def{def:label}.
\end{lemma}

\noindent\emph{Proof}. First assume $\bffirst(\vec{u}) \ne 0$, then $\vec{u} \in \mathcal{F}$ is the tensor product representation of a string term $t \in T_F$ labeling a leaf node in a tree. Therefore
\[
 \bflab(\gamma, \Psi(t)) = \Psi(t) = \Psi(\lab(\gamma, t)) \:.
\]
Next, suppose $\bffirst(\vec{u}) = 0$, in which case $\vec{u} \in \mathcal{F}$ is the tensor product representation of a proper tree term $t = f(t_0, t_1) \in T_A$. By means of induction over $\gamma$, we have first
\[
 \bflab(\varepsilon, \Psi(t)) =  (\id \otimes \vec{r}^+_2)(\vec{f} \otimes \vec{r}_2 \oplus \Psi(t_0) \otimes \vec{r}_0 \oplus \Psi(t_1) \otimes \vec{r}_1) =
 \vec{f} =\Psi(\lab(\varepsilon, t))
\]
and second
\begin{multline*}
 \bflab(i \gamma, \Psi(t)) = \bflab(\gamma, \bfext_i(\Psi(t))) =
 \bflab(\gamma, \Psi(\ext_i(t))) = \Psi(\lab(\gamma, \ext_i(t))) = \Psi(\lab(i \gamma, t)) \:.
\end{multline*}

The following function does not provide a Fock space realization but rather a kind of Fock space isometry.

\begin{definition}\label{def:tprhead}
The head of the tensor product representation of a minimalist tree $t \in T_A$ is obtained by a function $\bfhead:\mathcal{F}\to I$,
\[
    \bfhead(\vec{u}) = \left\{ \begin{array}{c@{\mbox{ if }}l}
                    \varepsilon & \bffirst(\vec{u}) \ne 0 \\
                    0^\frown \bfhead(\bfext_0(\vec{u})) & \vec{u} = \bfcons_<(\bfext_0(\vec{u}), \bfext_1(\vec{u})) \\
                    1^\frown \bfhead(\bfext_1(\vec{u})) & \vec{u} = \bfcons_>(\bfext_0(\vec{u}), \bfext_1(\vec{u})) \:.
            \end{array}
         \right.
\]
\end{definition}

\begin{lemma}\label{lem:tprhead}
Let $t \in T_A$. Then
\[
 \bfhead(\Psi(t)) = \head(t)
\]
with $\head$ given in \Def{def:head}.
\end{lemma}

\noindent\emph{Proof} (by induction). First assume that $\bffirst(\vec{u}) \ne 0$, i.e. $t \in T_F$ with $\vec{u} = \Psi(t)$ is a head. Then
\[
\bfhead(\Psi(t)) = \varepsilon = \head(t) \:.
\]
Next suppose $t = f(t_0, t_1) \in T_A$ with either $f = \mathtt{<}$ or $f = \mathtt{>}$ .  In the first case we have $\vec{u} = \bfcons_<(\bfext_0(\vec{u}), \bfext_1(\vec{u}))$ and therefore
\[
 \bfhead(\Psi(t)) = 0^\frown \bfhead(\bfext_0(\vec{u})) = 0^\frown \head(\ext_0(t)) \:,
\]
in the second case we have $\vec{u} = \bfcons_>(\bfext_0(\vec{u}), \bfext_1(\vec{u}))$ and thus
\[
 \bfhead(\Psi(t)) = 1^\frown \bfhead(\bfext_1(\vec{u})) = 1^\frown \head(\ext_1(t)) \:.
\]

\begin{definition}\label{def:tprfeature}
The feature of the tensor product representation $\vec{u} = \Psi(t)$ of a minimalist tree $t$ is retained as the first feature of $t$'s head label. Thus $\bffeat: \mathcal{F} \to \mathcal{V}_F$,
\[
 \bffeat(\vec{u}) = \bffirst(\bflab(\bfhead(\vec{u}), \vec{u}))  \:.
\]
\end{definition}

\begin{lemma}\label{lem:tprfeature}
Let $t \in T_A$. Then
\[
 \bffeat(\Psi(t)) = \Psi(\feat(t))
\]
with $\feat$ given in \Def{def:feature}.
\end{lemma}

\noindent\emph{Proof}. Follows immediately from previous lemmata and definitions.

Also the maximal projection becomes an analogue to a Fock space isometry.

\begin{definition}\label{def:tprmaxpro}
Let $\vec{u} \in \mathcal{F}$ and $\gamma \in I$. Then, $\bfmax: I \times \mathcal{F} \to I$,
\[
    \bfmax(\gamma, \vec{u}) = \left\{ \begin{array}{c@{\mbox{\quad:\quad}}l}
                         \varepsilon & \gamma = \bfhead(\vec{u}) \\
                         i^\frown \bfmax(\delta, \bfext_i(\vec{u})) &  \gamma = i \delta \mbox{ and } \gamma \ne \bfhead(\vec{u}) \\
                         \mbox{undefined} & \mbox{otherwise} \:.
                             \end{array}
         \right.
\]
\end{definition}

\begin{lemma}\label{lem:tprmaxpro}
Let $t \in T_A$ and $\gamma \in I$. Then
\[
 \bfmax(\gamma, \Psi(t)) = \max(\gamma, t)
\]
with $\max$ given in \Def{def:maxpro}.
\end{lemma}

\noindent\emph{Proof} (by induction over $\gamma$). Let $\gamma = \bfhead(\Psi(t))$. Then
\[
  \bfmax(\gamma, \Psi(t)) = \varepsilon = \max(\gamma, t) \:,
\]
if $\gamma \ne \bfhead(\Psi(t))$, by contrast, we find $\delta \in I$ such that $\gamma = i \delta$, hence
\[
  \bfmax(\gamma, \Psi(t)) = i^\frown \bfmax(\delta, \bfext_i(\Psi(t))) = i^\frown \max(\delta, \ext_i(t)) = \max(\gamma, t) \:.
\]

The function $\bfmax$ is naturally extended to sets of node addresses.
\begin{definition}\label{def:tprmaxpro2}
Let $\vec{u} \in \mathcal{F}$ and $P \subset I$. Then, $\bfmax^\#: \wp(I) \times \mathcal{F} \to \wp(I)$,
\[
    \bfmax^\#(P, \vec{u}) = \bigcup_{\gamma \in P} \{ \bfmax(\gamma, \vec{u}) \} \:.
\]
\end{definition}

Next we have to adapt the definition of the symbolic $\leav$ function from \Def{def:leaves}. The corresponding realization $\bfleav: \mathcal{V}_F \times \mathcal{F} \to \wp(I)$ is obtained from a generalized unbinding function
\begin{equation}\label{eq:tprleavub}
   \bfubfeat(\gamma, \vec{f}, \vec{u}) = (\vec{f}^+ \otimes \vec{s}_1^+ \otimes \vec{r}_{\gamma}^+)(\vec{u}) \:,
\end{equation}
for given filler vector $\vec{f} \in \mathcal{V}_F$ and leaf address $\gamma \in I$, applied to the tensor product representation of a tree $t \in T_A$, because all first features of the tree's leaves built partial sums of the form
\begin{equation}\label{eq:tprleav}
   \sum_{i=1}^m \vec{f}_i \otimes \vec{s}_1 \otimes \vec{r}_{\eta_i} \:,
\end{equation}
as they are bound to the first role $\vec{s}_1$ in the feature lists. Here, $\vec{r}_{\eta_i}$ denote the multiple tensor products of roles according to \Def{eq:roleaddr}.

Applying \Eq{eq:tprleavub} to this expression yields
\begin{multline*}
 \bfubfeat(\gamma, \vec{f}, \vec{u}) =
 (\vec{f}^+ \otimes \vec{s}_1^+ \otimes \vec{r}_{\gamma}^+) \left( \sum_{i=1}^m \vec{f}_i \otimes \vec{s}_1 \otimes \vec{r}_{\eta_i} \right) =
 \sum_{i=1}^m \vec{f}^+(\vec{f}_i) \,  \vec{s}_1^+(\vec{s}_1) \, \vec{r}_{\gamma}^+(\vec{r}_{\eta_i}) = \delta_{\gamma, {\eta_i}}
\end{multline*}
for all $\vec{f}_i = \vec{f}$.

Therefore we get
\begin{definition}\label{def:tprleaves}
Let $\vec{f} \in \mathcal{V}_F$ and $\vec{u} \in \mathcal{F}$ . Then, $\bfleav: \mathcal{V}_F \times \mathcal{F} \to \wp(I)$,
\[
  \bfleav(\vec{f}, \vec{u}) = \{ \gamma \in I | \bfubfeat(\gamma, \vec{f}, \vec{u}) = 1 \} \:.
\]
\end{definition}

\begin{lemma}\label{lem:tprleaves}
Let $t \in T_A$ and $f \in F_F$. Then
\[
 \bfleav(\Psi(f), \Psi(t)) = \leav(f, t) \:.
\]
\end{lemma}

\noindent\emph{Proof}. The lemma follows from the above calculation.

Next, we modify the replacement function.

\begin{definition}\label{def:tprreplace}
Let $\vec{u}, \vec{u}' \in \mathcal{F}$ and $\gamma \in I$. Then $\bfrep : I \times \mathcal{F} \times \mathcal{F} \to \mathcal{F}$ with
\begin{eqnarray*}
  \bfrep(\varepsilon, \vec{u}, \vec{u}') &=& \vec{u}' \\
  \bfrep(0 \gamma, \vec{u}, \vec{u}') &=& \bfcons_{\bflab(\varepsilon, \vec{u})}(\bfrep(\gamma, \bfext_0(\vec{u}), \vec{u}'), \bfext_1(\vec{u})) \\
  \bfrep(1 \gamma, \vec{u}, \vec{u}') &=& \bfcons_{\bflab(\varepsilon, \vec{u})}(\bfext_0(\vec{u}), \bfrep(\gamma, \bfext_1(\vec{u}), \vec{u}')) \:.
\end{eqnarray*}
\end{definition}

\begin{lemma}\label{lem:tprreplace}
Let $t, t' \in T_A$ and $\gamma \in I$.  Then
\[
 \bfrep(\gamma, \Psi(t), \Psi(t')) = \Psi(\rep(\gamma, t, t'))
\]
with $\rep$ from \Def{def:replace}.
\end{lemma}

\noindent\emph{Proof} (by means of induction over $\gamma$). First let $\gamma = \varepsilon$. Then
\[
 \bfrep(\varepsilon, \Psi(t), \Psi(t')) = \Psi(t) = \Psi(\rep(\varepsilon, t, t')) \:.
\]
Next assume that Lemma \ref{lem:tprreplace} has already been proven for all address strings $\gamma$ of length $p \in \mathbb{N}_0$. Then $i \gamma$ with $i = 0$ or $i = 1$ is of length $p+1$ and it holds either
\begin{multline*}
 \bfrep(0 \gamma, \Psi(t), \Psi(t')) = \bfcons_{\bflab(\varepsilon, \Psi(t))}(\bfrep(\gamma, \bfext_0(\Psi(t)), \Psi(t')), \bfext_1(\Psi(t))) = \\
 \bfcons_{\Psi(\lab(\varepsilon, t))}(\Psi(\rep(\gamma, \ext_0(t), t')), \Psi(\ext_1(t))) =
 \Psi(\cons_{\lab(\varepsilon, t)}(\rep(\gamma, t, t'), \ext_1(t))) = \\
 \Psi(\rep(\gamma, t, t'))
\end{multline*}
or
\begin{multline*}
 \bfrep(1 \gamma, \Psi(t), \Psi(t')) = \bfcons_{\bflab(\varepsilon, \Psi(t))}(\bfext_0(\Psi(t)), \bfrep(\gamma, \bfext_1(\Psi(t)), \Psi(t'))) = \\
 \bfcons_{\Psi(\lab(\varepsilon, t))}(\Psi(\ext_0(t)), \Psi(\rep(\gamma, \ext_1(t), t'))) = \\
 \Psi(\cons_{\lab(\varepsilon, t)}(\ext_0(t), \rep(\gamma, \ext_1(t), t'))) = \Psi(\rep(\gamma, t, t')) \:.
\end{multline*}

Using the Fock space realization of replace we also extend the domain of the shift function (\ref{def:tprstrfnc}) from string vectors in $\mathcal{S}$ to tree vectors in $\mathcal{F}$.
\begin{definition}\label{def:tprshift2}
Let $\vec{u} \in \mathcal{F}$. Then, $\bfshift^\#: \mathcal{F} \to \mathcal{F}$ with
\[
  \bfshift^\#(\vec{u}) = \bfrep(\bfhead(\vec{u}), \vec{u}, \bfshift(\bflab(\bfhead(\vec{u}), \vec{u}))) \:.
\]
\end{definition}

\begin{lemma}\label{lem:tprshift2}
Let $t \in T_A$.  Then
\[
 \bfshift^\#(\Psi(t)) = \Psi(\shift^\#(t))
\]
with $\shift^\#$ from \Def{def:shift2}.
\end{lemma}

\noindent\emph{Proof}. The Lemma follows from previous observations.

% ------------------------------------- Section -----------------------------------
\subsection{Minimalist grammars}
\label{sec:tprmg2}

In this section we introduce geometric minimalist structure-building functions and prove that they are indeed Fock space realizations of the term algebraic functions from \Sec{sec:mg2}.

\begin{definition}\label{def:tprmg}
Let $G = (P, C, \Lex, \mathcal{M})$ be a minimalist grammar (MG) with phonetic features $P$, categories $C = B \cup S \cup L \cup M$, lexicon $\Lex \subset T_A$, and structure-building functions $\mathcal{M} = \{ \merge, \move \}$ as defined in \Def{def:mg}. Let $\sel: S \to B$ be the select function and $\lic: L \to M$ be the license function. Moreover, let $\Psi = \psi \circ \beta_A$ be a tensor product representation of the term algebra $T_A$ of $G$ on Fock space $\mathcal{F}$. We introduce realizations $\bfsel: \mathcal{F} \to \mathcal{F}$ and $\bflic: \mathcal{F} \to \mathcal{F}$ by demanding
\begin{eqnarray*}
\Psi(\sel(s)) &=& \bfsel(\Psi(s)) \\
\Psi(\lic(\ell)) &=& \bflic(\Psi(\ell))
\end{eqnarray*}
for $s \in S$ and $\ell \in L$. The domain of $\bfmerge$ is given by all pairs of vectors $\bfDomMerge = \{ (\vec{u}_1, \vec{u}_2) \in \mathcal{F} \times \mathcal{F} | \bfsel(\bffeat(\vec{u}_1)) = \bffeat(\vec{u}_2) \}$. The domain of $\bfmove$ contains all vectors $\bfDomMove = \{ \vec{u} \in \mathcal{F} | \bffeat(\vec{u}) \in \Psi(L) \text{ and } \bfmax^\#(\bfleav(\bflic(\bffeat(\vec{u})), \vec{u}), \vec{u}) \text{ contains}\\
\text{ exactly one element} \}$. Let $\vec{u}_1, \vec{u}_2 \in \bfDomMerge$ and $\vec{u} \in \DomMove$, then
    \begin{eqnarray*}
      \bfmerge(\vec{u}_1, \vec{u}_2) &=& \left\{ \begin{array}{c@{\mbox{ if }}l}
                    \bfcons_<(\bfshift^\#(\vec{u}_1), \bfshift^\#(\vec{u}_2)) & \bffirst(\vec{u}_1) \ne 0 \\
                    \bfcons_>(\bfshift^\#(\vec{u}_1), \bfshift^\#(\vec{u}_2)) & \bffirst(\vec{u}_1) = 0
                                        \end{array}
         \right. \\
      \bfmove(\vec{u}) &=& \bfcons_>( \bfshift^\#(\bfext_{\bfmax(\bfleav(\bflic(\bffeat(\vec{u})), \vec{u}), \vec{u})}(\vec{u})), \\
        && \bfshift^\#(\bfrep(\bfmax(\bfleav(\bflic(\bffeat(\vec{u})), \vec{u}), \vec{u}), \vec{u}, \varepsilon)))
    \end{eqnarray*}
\end{definition}

\begin{theorem}\label{theor:mgtpr}
Let $T_A$ be the minimalist tree term algebra and $\Psi$ its tensor product representation in Fock space $\mathcal{F}$, as above. Let $t_1, t_2 \in \DomMerge$ and $t \in \DomMove$, then
    \begin{eqnarray*}
    \bfmerge(\Psi(t_1), \Psi(t_2) &=& \Psi(\merge(t_1, t_2)) \\
    \bfmove(\Psi(t)) &=& \Psi(\move(t))
    \end{eqnarray*}
    with $\merge, \move$ from \Def{def:mg}.
\end{theorem}

\noindent\emph{Proof}. The Theorem follows from the Lemmata in \Sec{sec:tpr}.

Taken together, we have proven that derivational minimalism \citep{Stabler97, StablerKeenan03, Michaelis01} can be realized by tensor product representations as a starting point for integrated connectionist/symbolic architectures \citep{SmolenskyLegendre06a, Smolensky06}.

% ------------------------------------- Section -----------------------------------
\subsection{Processing Algorithm}
\label{sec:pars2}

In order to realize a minimalist bottom-up processor as discussed in \Sec{sec:pars} in Fock space, we have to represent the processor's
state descriptions \citep{Stabler96}. This can be achieved through another filler/role binding by introducing new roles $p_1, p_2, \dots \in R$ for stack positions binding minimalist trees. Then the tensor product representation of a state description $w$ of length $m$ assumes the form
\begin{equation}\label{eq:tprstatedes}
    \vec{w} = \sum_{k=1}^m \vec{w}_k \otimes \vec{p}_k \:,
\end{equation}
where $\vec{w}_k$ are tensor product representations of minimalist trees.

The minimalist algorithm as defined in \Def{def:pars} becomes then realized by corresponding Fock space functions $\bfmerge^*$ and $\bfmove^*$.

\begin{definition}\label{def:trppars}
Let $T_A$ be the minimalist tree term algebra and $\Psi$ its tensor product representation in Fock space $\mathcal{F}$, as above. Furthermore, let $\mathcal{F}$ be augmented by the role vectors of a minimalist state description. We define
\begin{enumerate}
\item $\bfmerge^*: \mathcal{F} \to  \mathcal{F}$ with
\begin{equation*}
    \bfmerge^*(\vec{w}) = \sum_{k=1}^{m-2} (\id \otimes \vec{p}_k^+)(\vec{w}) \otimes \vec{p}_k \oplus
    \bfmerge((\id \otimes \vec{p}_{m-1}^+)(\vec{w}), (\id \otimes \vec{p}_m^+)(\vec{w})) \otimes \vec{p}_{m-1} \:.
\end{equation*}
\item $\bfmove^*: \mathcal{F} \to  \mathcal{F}$ with
\begin{equation*}
    \bfmove^*(\vec{w}) = \sum_{k=1}^{m-1} (\id \otimes \vec{p}_k^+)(\vec{w}) \otimes \vec{p}_k \oplus
    \bfmove((\id \otimes \vec{p}_m^+)(\vec{w})) \otimes \vec{p}_m \:.
\end{equation*}
\end{enumerate}
\end{definition}
In \Def{def:trppars} the adjoint vectors $\vec{p}_k^+$ applied to the tensor product representation $\vec{w}$ yield the corresponding expressions $\vec{w}_k$ from \Eq{eq:tprstatedes}. Clearly, this definition entails a minimalist processor as stated by the next theorem.

\begin{theorem}\label{theor:mgtprpars}
Let $T_A$ be the set of minimalist expressions and $\Psi$ the tensor product representation of its state descriptions in Fock space $\mathcal{F}$, as above. The functions $\bfmerge^*$ and $\bfmove^*$ given in \Def{def:trppars} realize a minimalist bottom-up processor in Fock space.
\end{theorem}

The proof of Theorem \ref{theor:mgtprpars} requires the realizability of permutation operators $\Pi: \mathcal{F} \to \mathcal{F}$ in Fock space. Such general permutations can be assembled from elementary transpositions $\tau_{ij}$, exchanging items $i$ and $j$ in an $m$-tuple. The corresponding realization $\mathbf{T}_{ij}$ is then obtained in the following way. Let
\[
    \vec{w} = \sum_{k=1}^m \vec{w}_k \otimes \vec{p}_k
\]
be the state description in Fock space and $\mathbf{P}_{ij}$ be the projector on the orthocomplement  spanned by $\vec{p}_i$ and $\vec{p}_j$ . Then
\begin{equation}\label{eq:transpo}
    \mathbf{T}_{ij}(\vec{w}) = \mathbf{P}_{ij}(\vec{w}) + (\id \otimes \vec{p}_i^+)(\vec{w}) \otimes \vec{p}_j + (\id \otimes \vec{p}_j^+)(\vec{w}) \otimes \vec{p}_i
\end{equation}
realizes the transposition $\tau_{ij}$ in Fock space $\mathcal{F}$ by means of unbinding functions. Then entries in the state description can be the rearrangement such that $\bfmerge^*$ and $\bfmove^*$ as defined in \Def{def:trppars} become applicable.

% ------------------------------------- Section -----------------------------------
\subsection{Harmonic minimalist grammars}
\label{sec:harmony}

A crucial component of ICS is \emph{harmony theory}. At the symbolic level of description, harmony assesses the well-formedness of a structure by means of \emph{soft-constraints} rewarding the minimization of markedness. It can be gauged in such a way, that totally well-formed output assumes harmony $H = 0$. By contrast, at the subsymbolic level of description, harmony provides a \emph{Lyapunov function} guiding the computational dynamics by means of \emph{gradient ascent}. In a neural network realization harmony of an activation vector $\vec{v}$ is given by a quadratic form
\[
    H(\vec{v}) = \vec{v}^+ \cdot \vec{W}(\vec{v}) \cdot \vec{v} \:,
\]
where $\vec{v}^+$ denotes the transposed of $\vec{v}$ and $\vec{W}(\vec{v})$ is the synaptic weight matrix in state $\vec{v}$ corresponding to the computational function applied to $\vec{v}$ \citep{Smolensky06, SmolenskyLegendre06a, SmolenskyLegendre06b}.

We owe a first indication of weighted or harmonic minimalist grammars (HMG) to \citet{Stabler97} who speculated about ``additional `economy principles,' acting as a kind of filter on derivations'' (see also \citet{Harkema01}). \Citet{Hale06} made the first attempt to implement this idea by constructing probabilistic context-free grammars from minimalist derivation trees. Therefore we suggest the following definition.

\begin{definition}\label{def:hmg}
A harmonic minimalist grammar (HMG) is a minimalist grammar $G$ (\Def{def:mg}) augmented with:
\begin{enumerate}
\item A weight function for feature terms $W: T_F \to \bigoplus_{p=1}^\infty \mathbb{R}^p$, such that $W(s)$ is a $p$-tuple $(x_1, x_2, \dots, x_p) \in \mathbb{R}^p$ of real weights assigned to a feature term $s = (f_1 \circ f_2 \circ \ldots \circ f_p)(\varepsilon)$ of length $p \in \mathbb{N}$. In particular, $W$ assigns weights to the features in the minimalist lexicon $\Lex$.
\item A harmony function for trees $H: T_A \to \mathbb{R}$, given by
\[
 H(t) = \vec{x}_1^+(W(\lab(\head(t), t))) \:,
\]
with the adjoint vector $\vec{x}_1^+$ of the direction of the $x_1$-axis: $\vec{x}_1^+(x_1, x_2, \dots, x_p) = x_1$, returns the weight of $t$'s head.
\item A collection of partial functions $\hmerge: \mathbb{R} \times T_A \times T_A \to \mathbb{R} \times T_A$ and $\hmove: \mathbb{R} \times T_A \to \mathbb{R} \times T_A$, defined as follows:
    \begin{eqnarray*}
      \hmerge(h, t_1, t_2) &=& (h + H(t_1) + H(t_2), \merge(t_1, t_2)) \\
      \hmove(h, t) &=& (h + H(t) + H(\ext_{\max(\leav(\lic(\feat(t)), t), t)}(t)), \move(t)) \:,
    \end{eqnarray*}
    for $h \in \mathbb{R}$.
\item The \emph{harmony filter}: A minimalist tree $t \in T_A$ is \emph{harmonically well-formed} if it is MG well-formed and additionally
\[
 h(t) \ge 0 \:,
\]
where $h(t)$ is the cumulative harmony of $t$ after application of $\hmerge$ and $\hmove$ during the derivation of $t$, starting with initial condition $h_0 \in \mathbb{R}$.
\end{enumerate}
\end{definition}

Next, we suggest a metric for geometric representations that can the regarded as a measure of harmony. For that aim we assume that the Fock space $\mathcal{F}$ is equipped with a norm $||\cdot|| : \mathcal{F} \to \mathbb{R}_0^+$ assigning a length $||\vec{u}||$ to vector $\vec{u} \in \mathcal{F}$. Such a norm could be supplied by a scalar product, when $\mathcal{F}$ is a Hilbert space.

\begin{definition}\label{def:harmony}
Let $(\vec{w}_1, \vec{w}_2, \dots, \vec{w}_T)$, $\vec{w}_k \in \mathcal{F}$, $1 \le k \le T$, $T \in \mathbb{N}$ be a (finite) trajectory in Fock space of duration $T$, representing a minimalist derivation with initial state $\vec{w}_1$ and final state $\vec{w}_T$. We define \emph{harmony} through the distance of an intermediate step $\vec{w}_k$ from the well-formed parse goal $\vec{w}_T$, i.e.
\[
    H(\vec{w}_k) = -||\vec{w}_k - \vec{w}_T|| \:.
\]
\end{definition}

\begin{lemma}\label{lem:harmony}
The harmony function from \Def{def:harmony} is non-positive for all processing steps and increases towards $H = 0$ when approaching the final state, $H(\vec{w}_T) = 0$.
\end{lemma}

\noindent\emph{Proof}. The Lemma follows immediately from \Def{def:harmony}.

Eventually we combine \Def{def:hmg} and \Def{def:harmony} by looking at harmony differences $\Delta H_k = H(\vec{w}_{k+1}) - H(\vec{w}_k)$ between successive parse steps. These differences can be distributed among the features triggering the transition from $\vec{w}_k$ to $\vec{w}_{k+1}$, as will be demonstrated in \Sec{sec:results}. HMG could then possibly account for gradience effects in language processing.

% ------------------------------------- Section -----------------------------------
\section{Applications}
\label{sec:appl}

In this section we present two example applications which use the tensor product representations of \Sec{sec:tpr} in different ways. Both representations are given here, since it is the aim of this paper to give theoretical justifications for both at the same time. The representations are using two different encodings. At first we show arithmetic representations implemented by \citet{Gerth06}, then, we describe fractal representations outlined by \citet{GerthGraben09b}. For computing harmony we use Euclidian norm in both cases.

% ------------------------------------- Section -----------------------------------
\subsection{Arithmetic Representation}
\label{sec:local}

In a first step, we map the fillers $F$ for the features of the lexical items onto 12 filler vectors as shown in \Tab{tab:tensors}.
\begin{table}[h]
  \centering
\begin{tabular}{ll}
    \hline
    \hline
  $\mathtt{d}$ & $\vec{f}_{1}$ \\
  $\mathtt{=d}$ & $\vec{f}_{2}$ \\
  $\mathtt{v}$ & $\vec{f}_{3}$ \\
  $\mathtt{=v}$ & $\vec{f}_{4}$ \\
  $\mathtt{t}$ & $\vec{f}_{5}$ \\
  $\mathtt{=t}$ & $\vec{f}_{6}$ \\
  $\mathtt{+CASE}$ & $\vec{f}_{7}$ \\
  $\mathtt{-case}$ & $\vec{f}_{8}$ \\
  $\mathtt{+I}$ & $\vec{f}_{9}$ \\
  $\mathtt{-i}$ & $\vec{f}_{10}$ \\
  $>$ & $\vec{f}_{11}$ \\
  $<$ & $\vec{f}_{12}$ \\
      \hline
\end{tabular}
  \caption{Fillers for the minimalist lexicon outlined in \Fig{fig:mg}.}\label{tab:tensors}
\end{table}

In order to ensure a faithful representation, filler vectors need to be linearly independent, i.e., they form a basis of 12-dimensional vector space. Trying to implement this requirement, leads to an explosion of dimensions (more than 5 millions) which was beyond the limits of memory on the used workstation. Therefore, we refrained from linear independence and used a linearly dependent, distributed, representation of filler vectors in a 4-dimensional vector space $\vec{f}_i \in \mathbb{R}^{4}$, $(1 \le i \le 12)$ instead.

The actual filler vectors are:
\[
\vec{f}_{1} = \left(\begin{array}{c}1 \\ 0 \\ 0 \\ 0\end{array}\right),
\vec{f}_{2} = \left(\begin{array}{c}0 \\ 1 \\ 0 \\ 0 \end{array}\right),
\vec{f}_{3} = \left(\begin{array}{c}0 \\ 0 \\ 1 \\Ê0\end{array}\right),
\vec{f}_{4} = \left(\begin{array}{c}0 \\ 0 \\ 0 \\ 1\end{array}\right),
\]
\[
\vec{f}_{5} = \frac{1}{\sqrt{3}} \left(\begin{array}{c}1 \\ 1 \\ 1 \\ 1\end{array}\right),
\vec{f}_{6} = \frac{1}{\sqrt{3}} \left(\begin{array}{c}-1 \\ 1 \\ 1 \\ 1\end{array}\right),
\vec{f}_{7} = \frac{1}{\sqrt{3}} \left(\begin{array}{c}1 \\ -1 \\ 1 \\ 1\end{array}\right),
\vec{f}_{8} = \frac{1}{\sqrt{3}} \left(\begin{array}{c}1 \\ 1 \\ -1 \\ 1\end{array}\right),
\]
\[
\vec{f}_{9} = \frac{1}{\sqrt{3}} \left(\begin{array}{c}1 \\ 1 \\ 1 \\ -1\end{array}\right),
\vec{f}_{10} = \frac{1}{\sqrt{3}} \left(\begin{array}{c}-1 \\ -1 \\ 1 \\ 1\end{array}\right),
\vec{f}_{11} = \frac{1}{\sqrt{3}} \left(\begin{array}{c}1 \\ -1 \\ -1\\ 1\end{array}\right),
\vec{f}_{12} = \frac{1}{\sqrt{3}} \left(\begin{array}{c} 1 \\ 1 \\ -1 \\ -1\end{array}\right).
\]

Similarly, the tree roles from \Fig{fig:treeroles} are represented by three-dimensional basis vectors as achieved in previous work \citep{GrabenGerthVasishth08, GerthGraben09b}. Further, we need to map the list positions  $s_i$ ($1 \le i \le 4$) of the features onto role vectors. Therefore, a total of $3+4=7$  role vectors is required. Again we have to use a linearly dependent representation for role vectors because of an explosion of dimensions and a restriction on available workstation memory.

In particular, we make the following assignment for tree roles ``left-daughter'' $\vec{r}_0 = \vec{e}_1$; ``right-daughter'' $\vec{r}_1 = \vec{e}_2$; ``mother'' $\vec{r}_2 = \vec{e}_3$, where $\vec{e}_k$ (k = 1, 2, 3) are the canonical basis vectors of three-dimensional space $\mathbb{R}^3$. The roles of list positions in the feature arrays of the minimalist lexicon  $\vec{r}_{i+2} = \vec{s}_i$ ($1 \le i \le 4$) are indicated in \Fig{fig:roles}.

\begin{figure}[H]
\[
\begin{bmatrix}
  \tt{=t} & \vec{r}_{3} \\ \mbox{c} \\
\end{bmatrix}
\begin{bmatrix}
  \tt{d} & \vec{r}_{3} \\  \tt{-case} & \vec{r}_{4} \\  \mbox{Douglas} \\
\end{bmatrix}
\begin{bmatrix}
  \tt{=d} & \vec{r}_{3}\\ \tt{v} & \vec{r}_{4} \\ \tt{-i} & \vec{r}_{5} \\ \mbox{love} \\
\end{bmatrix}
\begin{bmatrix}
  \tt{=v} & \vec{r}_{3}\\ \tt{+CASE} & \vec{r}_{4} \\ \tt{=d} & \vec{r}_{5} \\ \tt{v} & \vec{r}_{6} \\ \mbox{$\epsilon$} \\
\end{bmatrix}
\begin{bmatrix}
  \tt{=v} & \vec{r}_{3}\\ \tt{+I} & \vec{r}_{4}  \\ \tt{+CASE} & \vec{r}_{5} \\ \tt{t} & \vec{r}_{6} \\ \mbox{-ed} \\
\end{bmatrix}
\begin{bmatrix}
  \tt{d} & \vec{r}_{3}\\ \tt{-case} & \vec{r}_{4}\\ \mbox{deadlines} \\
\end{bmatrix}
\]
\caption{Roles for the Minimalist lexicon outlined in \Fig{fig:mg}.}\label{fig:roles}
\end{figure}

The vectors for the list positions are distributed on the unit sphere in $\mathbb{R}^{3}$:

\[
\vec{r}_{3} = \frac{1}{\sqrt{3}} \left(\begin{array}{c} 1 \\ 1 \\ 1\end{array}\right),
\vec{r}_{4} = \frac{1}{\sqrt{3}} \left(\begin{array}{c} -1 \\ 1 \\ 1\end{array}\right),
\vec{r}_{5} = \frac{1}{\sqrt{3}} \left(\begin{array}{c} 1 \\ -1 \\ 1\end{array}\right),
\vec{r}_{6} = \frac{1}{\sqrt{3}} \left(\begin{array}{c} 1 \\ 1 \\ -1\end{array}\right).
\]

The following example shows a tensor product representation of the lexical item for ``love":
\begin{figure}[H]
\[
\begin{bmatrix}
  \tt{=d} & \vec{f}_{2} \\ \tt{v} & \vec{f}_{3} \\  \tt{-i} & \vec{f}_{10} \\  \mbox{love}
\end{bmatrix} \otimes
\begin{bmatrix}
  \tt{=d} & \vec{r}_{3} \\ \tt{v} & \vec{r}_{4} \\ \tt{-i} & \vec{r}_{5} \\ \mbox{love}
\end{bmatrix}
=
\vec{f}_{2} \otimes \vec{r}_{3} \oplus \vec{f}_{3} \otimes \vec{r}_{4} \oplus \vec{f}_{10} \otimes \vec{r}_{5}
\]
\caption{Tensor product representation of the lexical item ``love".}\label{fig:tpr_lexitem}
\end{figure}

In our arithmetic tensor product representation, tensor products are then given as \textit{Kronecker products} \citep{Mizraji92} of filler and role vectors, $\vec{f}_i \otimes \vec{r}_k$, by:
\[
 \begin{pmatrix}
  f_1  \\
  f_2  \\
  \vdots \\
  f_{12}
 \end{pmatrix}
\otimes
\begin{pmatrix}
  r_0  \\
  r_1  \\
  \vdots \\
  r_6
 \end{pmatrix}
 =
 \begin{pmatrix}
    f_1 \begin{pmatrix} r_0  \\  r_1  \\  \vdots \\  r_6 \end{pmatrix} \\
    f_2 \begin{pmatrix} r_0  \\  r_1  \\  \vdots \\  r_6 \end{pmatrix} \\
    \vdots \\
    f_{12} \begin{pmatrix} r_0  \\  r_1  \\  \vdots \\  r_6  \end{pmatrix}
    \end{pmatrix}
     =
\begin{pmatrix}
    f_1 r_0  \\
    f_1 r_1  \\
    f_1 r_2  \\
    f_1 r_3  \\
    f_1 r_4  \\
    f_1 r_5  \\
    f_1 r_6  \\
    f_2 r_0  \\
    f_2 r_1  \\
    f_2 r_2  \\
    f_2 r_3  \\
    f_2 r_4  \\
    f_2 r_5  \\
    f_2 r_6  \\
  \vdots \\
    f_{12} r_0  \\
    f_{12} r_1  \\
    f_{12} r_2  \\
    f_{12} r_3 \\
    f_{12} r_4  \\
    f_{12} r_5  \\
    f_{12} r_6
 \end{pmatrix} \:.
\]

In order to construct an appropriate embedding space, we chose the largest tree appearing in the minimalist state description. The tensor product representation of every tree $t \in T_A$ is then embedded into that space by left-multiplication of the tree-roles with sufficient tensor powers
\[
 \vec{r}_2^{\otimes p} = \vec{r}_2 \otimes \vec{r}_2 \otimes \cdots \otimes \vec{r}_2
\]
($p$ times) of the mother role, where the exponent $p \in \mathbb{N}_0$ is different for every tree.

Finally, we have to construct the tensor product representation for the state descriptions of a minimalist bottom-up processor as described in \Sec{sec:pars2}. Here, we bind all minimalist expressions to only one role $p_0$ for the state description. For the tensor product representation, we simply choose $\vec{p}_0 = 1$, i.e. the scalar unit. As a result, all tree representing vectors become linearly superimposed in the state description \citep{SmolenskyLegendre06a}.

% ------------------------------------- Section -----------------------------------
\subsection{Fractal Tensor Product Representation}
\label{sec:frac}

\Citet{GerthGraben09b} introduced a different encoding called \emph{fractal tensor product representation} which is a combination of the arithmetic description in the previous section and scalar G\"odel encodings \citep{GrabenPotthast09a, GerthGraben09b}. For a fractal representation we encode the three tree roles $r_0, r_1, r_2$ localistically by the canonical basis vectors of three-dimensional vector space as above. However, fillers for minimalist features are represented by integer numbers $g(f_i)$ from a G\"odel encoding. The G\"odel codes used in our example are shown in \Tab{tab:fractals}.

\begin{table}[H]
  \centering
\begin{tabular}{ll}
    \hline
  Filler $f_i$ & Code \\
    \hline
  $\mathtt{d}$ & 0 \\
  $\mathtt{=d}$ & 1 \\
  $\mathtt{v}$ & 2 \\
  $\mathtt{=v}$ & 3 \\
  $\mathtt{t}$ & 4 \\
  $\mathtt{=t}$ & 5 \\
  $\mathtt{+CASE}$ & 6 \\
  $\mathtt{-case}$ & 7 \\
  $\mathtt{+I}$ & 8 \\
  $\mathtt{-i}$ & 9 \\
   $>$ & 10 \\
  $<$ & 11 \\
      \hline
\end{tabular}
  \caption{Fractal encoding for minimalist lexicon in \Fig{fig:mg}.}\label{tab:fractals}
\end{table}

The role vectors of the tree positions are mapped onto three-dimensional vectors in the same way as described in \Sec{sec:local}. The only difference is the encoding of the positions of the lexical items in the feature array. Here, the roles $s_k$ are encoded by fractional powers $N^{-k}$ of the total number of fillers, which is $N=12$ and $k$ denotes the $k$-th list position. The following example shows the lexical entry for ``love" and its fillers represented as G\"odel numbers:
\[
L_{love} = \begin{bmatrix}
  \tt{=d} & 1 \\ \tt{v} & 2 \\  \tt{-i} & 9 \\ \mbox{love} \\
\end{bmatrix},
\]

It becomes described by the sum of (tensor) products of G\"odel numbers for the fillers and fractions for the list positions:
\[
 g(L_{love}) = 1 \times 12^{-1} + 2 \times 12^{-2} + 9 \times 12^{-3}= 0.1024 \:.
\]

The next example illustrates the encoding of a subtree, consider the tree:
\begin{figure}[H]
\centering
\Tree [.$<$ [ {$\begin{bmatrix} \tt{v} \\ \tt{-i} \\ \mbox{love} \\ \end{bmatrix}$} ]
		  [ {$\begin{bmatrix} \tt{-case} \\ \mbox{deadlines} \\ \end{bmatrix}$} ]
	 ].$<$
\end{figure}
Its encoding is given through
\begin{multline}
g(<) \otimes \vec{r}_2 \oplus g(L_l) \vec{r}_0 \oplus g(L_r) \otimes \vec{r}_1 \\
= 11 \times 12^{-1} \left(\begin{array}{c} 0 \\0 \\1 \end{array}\right) + (2 \times 12^{-1} + 9 \times 12^{-2}) \left(\begin{array}{c} 1 \\0 \\0\end{array}\right) + 7 \times 12^{-1} \left(\begin{array}{c} 0 \\1 \\0\end{array}\right) \\
= \left(\begin{array}{c} 0.229 \\0.583 \\ 0.917\end{array}\right) \:,
\end{multline}
where $L_l$ and $L_r$ denote the feature arrays of the left and right leaf.
Complex trees are again represented by Kronecker products (see \Sec{sec:local} for details).

The state description of the algorithm is mapped step by step onto the fractal tensor product representation. At first, each leaf in the tree is encoded in an enumeration of fractals. In the second step the encoding of the whole state description is achieved by recursively binding minimalist trees as complex fillers to 3-dimensional role vectors. Finally the representation of all trees in the state description is linearly superimposed in a suitable embedding space.

% ------------------------------------- Section -----------------------------------
\subsection{Results}
\label{sec:results}

In this section we present the results of the applications obtained in the previous sections (\Sec{sec:local}, \Sec{sec:frac}).

The final derivation of the minimalist algorithm (\Sec{sec:applMG}) results in a matrix which is the state space trajectory. Each column stands for one derivational step in form of a vector in a high-dimensional embedding space. The dimensions of the final embedding space are $d = 78732$ for the arithmetic representation and $d = 6561$ for the fractal tensor product representation.

For visualization purposes the data have to be compressed. A common technique in multivariate statistics is the principal component analysis (PCA), which has been used as an observable model previously \citep{GrabenGerthVasishth08, GerthGraben09b}. Before applying the PCA the trajectories are standardized using \textit{z-}transformation to obtain a transformed distribution with zero mean and unit variance. Then the greatest variance in the data is in the direction of the first principal component, the second greatest variance is in the direction of the second principal component and so on. Plotting the first, PC\#1, and the second, PC\#2, principal component as observables against each other, entails a two-dimensional \emph{phase portrait} as an appropriate visualization of the processing geometry.

\begin{figure}[H]
\centering
 \subfigure[]{\includegraphics[width=0.45\textwidth]{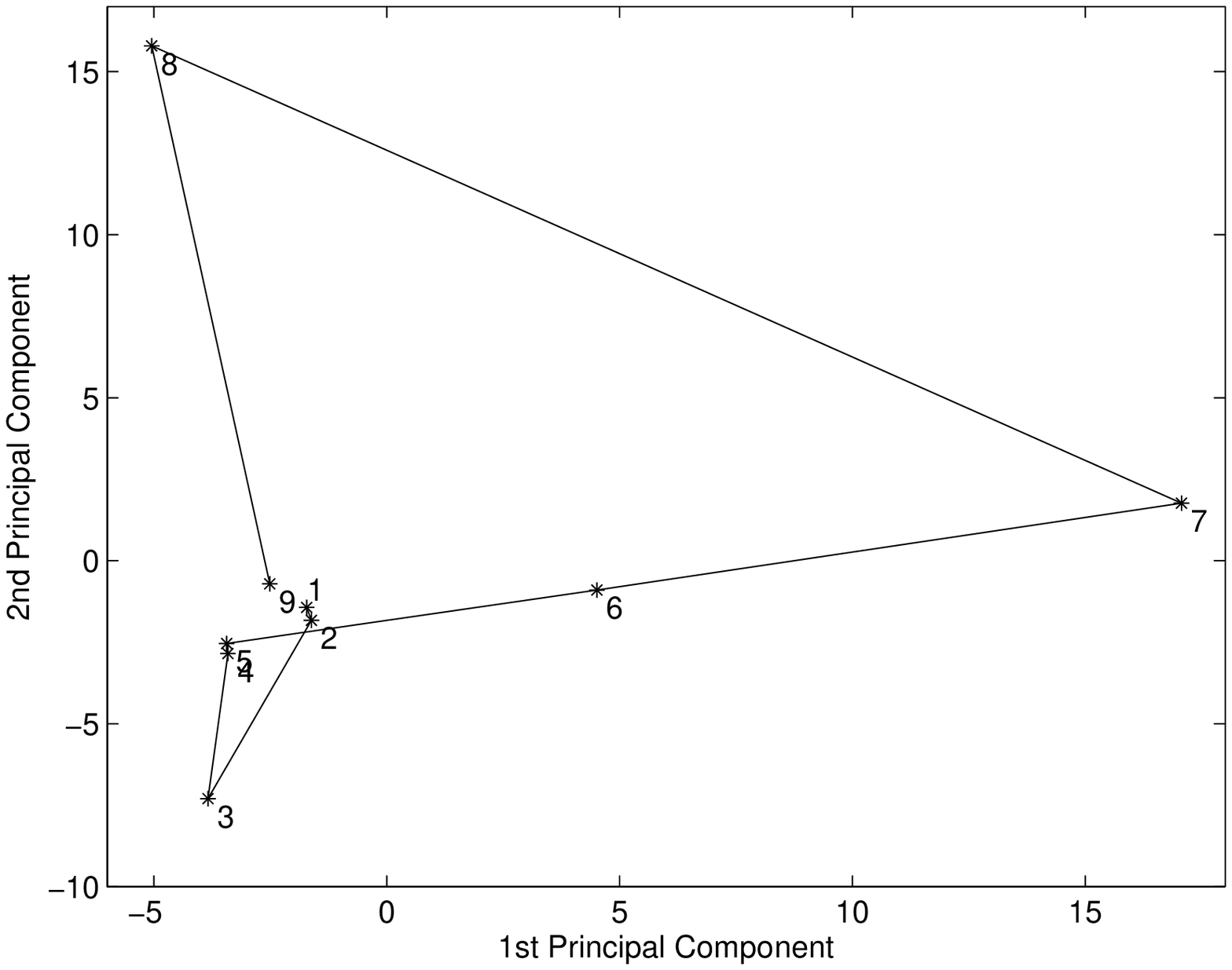}}
 \subfigure[]{\includegraphics[width=0.45\textwidth]{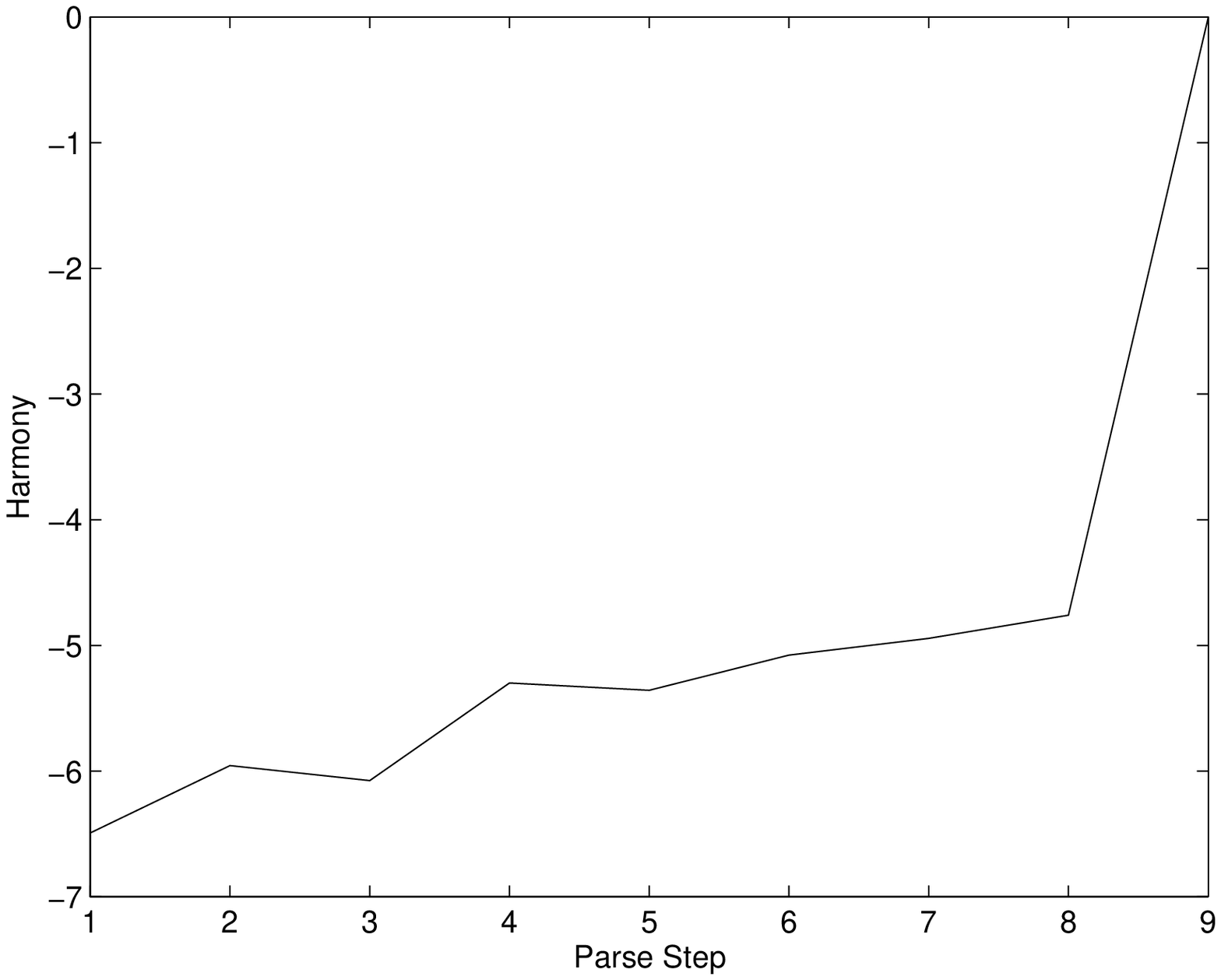}}
 \caption{\label{fig:pp_tensor} Results for the arithmetic representation (\Sec{sec:local}). (a) Phase portrait of the first principal component, PC\#1, versus the second principal component, PC\#2. (b) Harmony time series from \Def{def:harmony}. }
\end{figure}

First, we present the phase portrait and the harmony time series from \Def{def:harmony} of the arithmetic representation for sentence \ref{ex:adams} in \Sec{sec:applMG} in \Fig{fig:pp_tensor}.

Figure \ref{fig:pp_tensor}(a) shows the phase portrait in principal component space. Each parse step is subsequently numbered. Figure \ref{fig:pp_tensor}(b) presents the temporal development of the harmony function.

The derivation unfolds as described in \Sec{sec:applMG}. The initial state description (step 1) represents the lexicon and starts in coordinate $(-1.72, -1.43)$ in \Fig{fig:pp_tensor}(a) with a harmony value of $H = -6.49$ [\Fig{fig:pp_tensor}(b)]. As the parse continues the harmony trajectory climbs steadily upwards. In parse step 3 $\varepsilon$ is merged to the tree [\Fig{fig:pp_tensor}(a)]: coordinate $(-3.83, -7.3)$). Interestingly the graph of the harmony reaches a local minimum in $H = -6.08$ here and continues again upwards until parse step 8 [\Fig{fig:pp_tensor}(a)]: coordinate $(-5.05, 15.70)$; \Fig{fig:pp_tensor}(b): $H = -4.76$. In this step the subject ``Douglas" is moved upwards leading to the final phonetic, but not yet fully syntactically parsed, representation of the sentence. In the end the graphs reach their final states in coordinate $(-2.51, -0.71)$[\Fig{fig:pp_tensor}(a)] and in $H = 0$ [\Fig{fig:pp_tensor}(b)].

\begin{figure}[H]
\centering
 \subfigure[]{\includegraphics[width=0.45\textwidth]{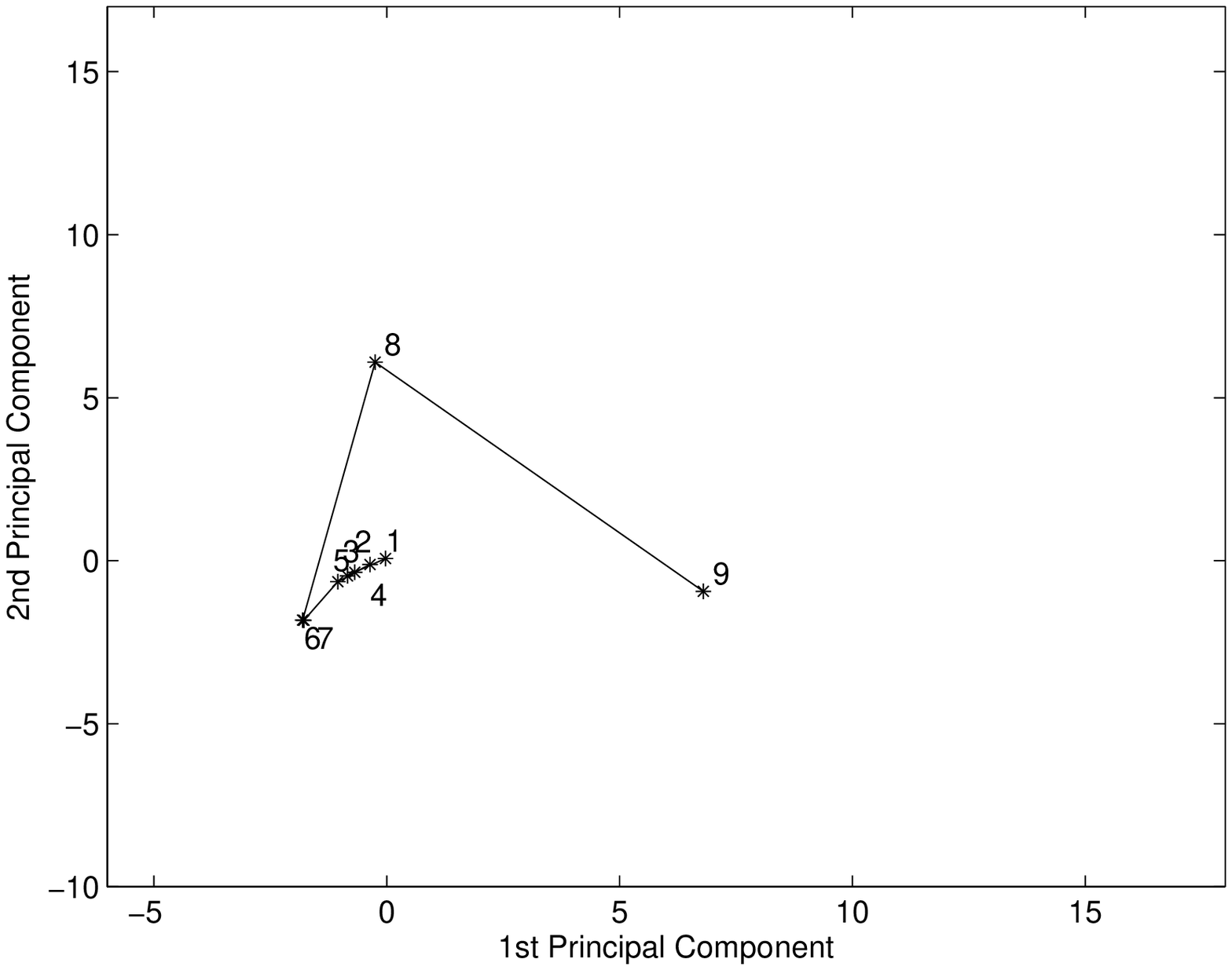}}
 \subfigure[]{\includegraphics[width=0.45\textwidth]{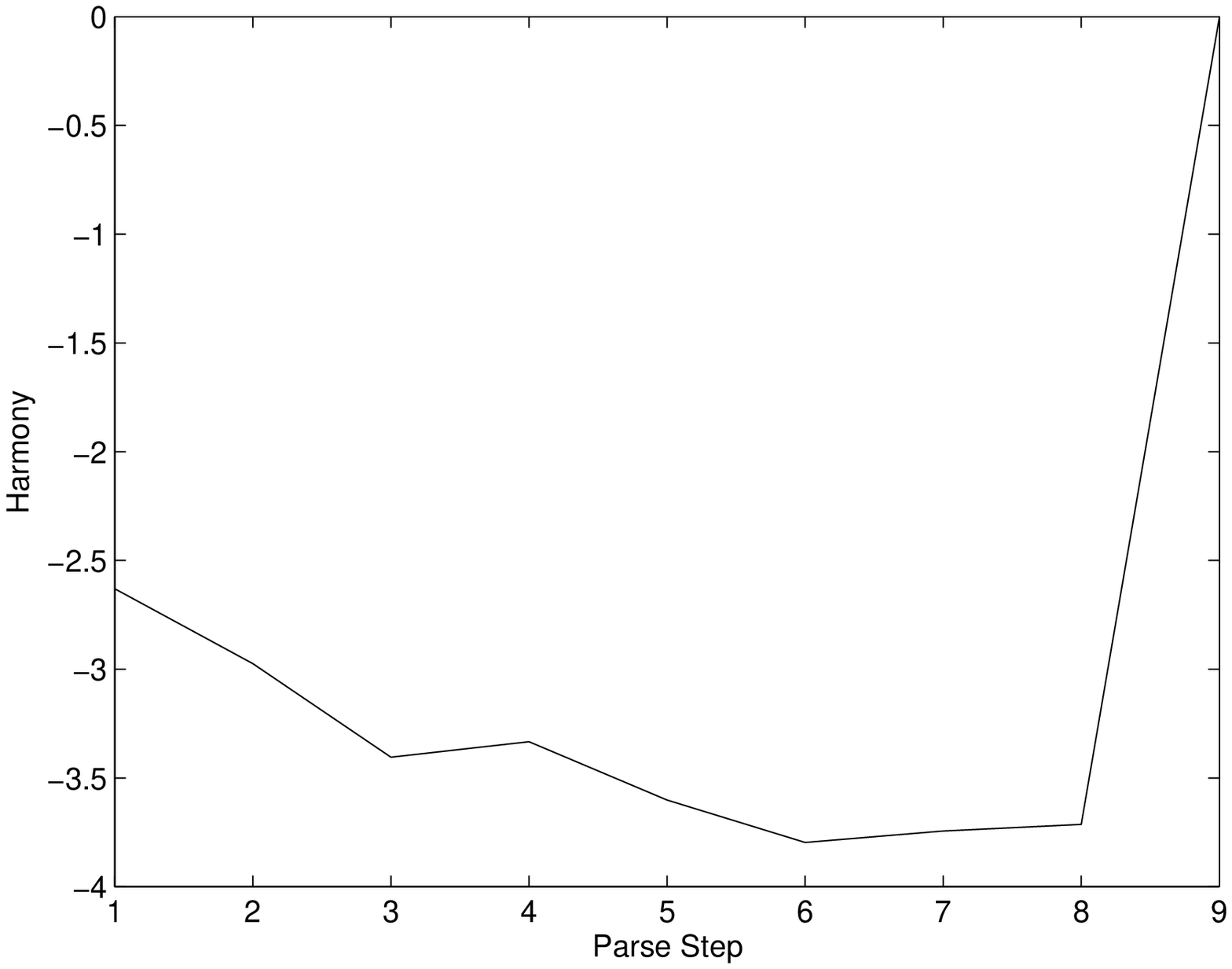}}
\caption{\label{fig:pp_fractal} Results for the fractal representation (\Sec{sec:frac}). (a) Phase portrait of the first principal component, PC\#1, versus the second principal component, PC\#2. (b) Harmony time series from \Def{def:harmony}. }
\end{figure}

Figure \ref{fig:pp_fractal} shows the observables for the processing mapped onto the fractal representation. Figure \ref{fig:pp_fractal}(a) displays the phase portrait in principal component space. Besides the apparent nonlinearity, one realizes another interesting property of the fractal representation: While the minimalist processing unfolds, the feature arrays contract. This is reflected by the increasing phase space volume available to the geometric dynamics. As above, \Fig{fig:pp_fractal}(b) illustrates the temporal development of the harmony function. Again, the initial state description represents all entries in the lexicon which starts in coordinate $(-0.03, 0.07)$ in \Fig{fig:pp_fractal}(a) with a harmony value of $H = -2.63$ in \Fig{fig:pp_fractal}(b). In comparison to \Fig{fig:pp_tensor}(a) the representations of the first seven parse steps stay close to each other before deviating to coordinate $(-0.25, 6.09)$ in step 8. The harmony curve in figure \Fig{fig:pp_fractal}(b) exhibits a downwards trend. Like in \Fig{fig:pp_tensor}(b) the graph of the harmony reaches a local minimum in parse step 3 ($H=-3.4$) when $\varepsilon$ is merged to the tree [\Fig{fig:pp_fractal}(a)]: coordinate $(-7.3, -5,8)$. Finally the end states are reached in coordinate $(6.8, -0.94)$ [\Fig{fig:pp_fractal}(a)] and in a harmony value of $H = 0$ [\Fig{fig:pp_fractal}(b)].

Table \ref{tab:harmonies} summarizes the evolution of harmonies for both representations.
\begin{table}[H]
  \centering
\begin{tabular}{*{11}{r}}
    \hline
  Representation & Step: 1 & 2 & 3 & 4 & 5 & 6 & 7 & 8 & 9 \\
    \hline
    arithmetic (\Sec{sec:local}) & $-6.49$  & $-5.96$ &  $-6.08$  & $-5.3$  & $-5.36$ &  $-5.08$  & $-4.94$ & $-4.76$ & $0$ \\
    fractal (\Sec{sec:frac}) &  $-2.63$  & $-2.97$ & $-3.4$&  $-3.33$  & $-3.6$ &  $-3.8$  & $-3.74$  & $-3.71$ & $0$ \\
        \hline
\end{tabular}
  \caption{\label{tab:harmonies} Harmony time series for both tensor product representations.}
\end{table}

Finally, we construct HMGs from these data by assigning harmony differences to the features of the minimalist lexicon as follows: First, we compute harmony differences $\Delta H_k = H(\vec{w}_{k+1}) - H(\vec{w}_k)$ between successive processing steps from \Tab{tab:harmonies}. Then, the difference $\Delta H_k$ is assigned to either a selector or a licensor that triggers the transition from $\vec{w}_k$ to $\vec{w}_{k+1}$ while the corresponding basic categories or licensees are weighted with zero.

Figure \ref{fig:hmg1} depicts the resulting HMG lexicon for the arithmetic representation (\Sec{sec:local}).
\begin{figure}[H]
\[
\begin{bmatrix}
  \tt{=t} & 4.76 \\ \mbox{c} \\
\end{bmatrix}
\begin{bmatrix}
  \tt{d} & 0 \\  \tt{-case} & 0 \\  \mbox{Douglas} \\
\end{bmatrix}
\begin{bmatrix}
  \tt{=d} & 0.54 \\ \tt{v} & 0 \\ \tt{-i} & 0 \\ \mbox{love} \\
\end{bmatrix}
\begin{bmatrix}
  \tt{=v} & -0.12 \\ \tt{+CASE} & 0.77 \\ \tt{=d} & -0.06 \\ \tt{v} & 0 \\ \mbox{$\epsilon$} \\
\end{bmatrix}
\begin{bmatrix}
  \tt{=v} & 0.28 \\ \tt{+I} & 0.13 \\ \tt{+CASE} & 0.18 \\ \tt{t} & 0 \\ \mbox{-ed} \\
\end{bmatrix}
\begin{bmatrix}
  \tt{d} & 0 \\ \tt{-case} & 0 \\ \mbox{deadlines} \\
\end{bmatrix}
\]
\caption{Harmonic minimalist lexicon of sentence \ref{ex:adams} obtained from arithmetic representation (\Sec{sec:local}).}\label{fig:hmg1}
\end{figure}

Moreover, \Fig{fig:hmg2} shows the HMG lexicon for the fractal representation (\Sec{sec:frac}).
\begin{figure}[H]
\[
\begin{bmatrix}
  \tt{=t} & 3.71 \\ \mbox{c} \\
\end{bmatrix}
\begin{bmatrix}
  \tt{d} & 0 \\  \tt{-case} & 0 \\  \mbox{Douglas} \\
\end{bmatrix}
\begin{bmatrix}
  \tt{=d} & -0.34 \\ \tt{v} & 0 \\ \tt{-i} & 0 \\ \mbox{love} \\
\end{bmatrix}
\begin{bmatrix}
  \tt{=v} & -0.43 \\ \tt{+CASE} & 0.0712 \\ \tt{=d} & -0.27 \\ \tt{v} & 0 \\ \mbox{$\epsilon$} \\
\end{bmatrix}
\begin{bmatrix}
  \tt{=v} & -0.19 \\ \tt{+I} & 0.05 \\ \tt{+CASE} & 0.03 \\ \tt{t} & 0 \\ \mbox{-ed} \\
\end{bmatrix}
\begin{bmatrix}
  \tt{d} & 0 \\ \tt{-case} & 0 \\ \mbox{deadlines} \\
\end{bmatrix}
\]
\caption{Harmonic minimalist lexicon of sentence \ref{ex:adams} obtained from fractal representation (\Sec{sec:frac}).}\label{fig:hmg2}
\end{figure}

% ------------------------------------- Section -----------------------------------
\section{Discussion}
\label{sec:disc}

In this paper we developed a geometric representation theory for minimalist grammars (MG). We resumed minimalist grammars in terms of partial functions acting on term algebras of trees and feature arrays. Those complex data structures were mapped onto vectors in a geometric space (known as the Fock space \citep{Haag92, SmolenskyLegendre06a}) using filler/role bindings and tensor product representations \citep{SmolenskyLegendre06a, Smolensky06, GrabenPotthast09a}. We were able to prove that the minimalist structure-building functions merge and move can be realized as piecewise linear maps upon geometric vector spaces. In order to present a proof-of-concept, we generalized the merge and move functions towards state descriptions of a simple derivation procedure for minimalist trees which also found a suitable realization in representation space. In addition, we suggested a harmony function measuring the distance of an intermediate processing state from a well-formed final state in representation space that gave rise to an extension of MG towards harmonic MG (HMG). This harmony observable could be regarded as a metric for processing complexity. While our proofs essentially relied on faithful representations, we used two different kinds of non-faithful, distributed representations in our numerical applications. Firstly, we employed arithmetic vector space encodings of minimalist features, roles and trees. Secondly, we used fractal tensor product representations that combine arithmetic vector spaces with numeric G\"odel encodings. For both cases, we presented phase portraits in principal component space and harmony time series of the resulting minimalist derivations. Finally, we derived the corresponding HMGs from simulated harmony differences.

Our theory proves that sophisticated grammar formalisms such as MG can be realized in a geometric representation. This would be a first step for dynamic cognitive modeling of an integrated connectionist/symbolic (ICS/DCM) architecture for processing minimalist grammars. Since natural languages tentatively belong to the same complexity class of mildly context-sensitive languages \citep{Shieber85, Stabler04}, ICS/DCM architectures are principally able to process natural language. However, the simple processing algorithm used in the present study just for illustrating the representation theory, is not a sound and complete minimalist parser \citep{Harkema01, Mainguy10, Stabler11b}. Therefore, future work towards psycholinguistically more plausible processing models, would comprise the development of a geometric representation theory for chain-based minimalism and for multiple context-free parsing \citep{Harkema01, StablerKeenan03}.

Moreover, processing minimalist grammars by ICS/DCM architectures straightforwardly provides a notion of harmony. However, a proper treatment of HMG would require further investigations to be carried out: Our definition of harmony in \Def{def:harmony} combines a particular metric (e.g. Euclidian) with one well-formed reference state $\vec{w}_T$ for minimalist processing, while harmony in ICS is defined as a general quadratic form only depending on the synaptic weight matrix. Therefore, one has to examine how these expressions would transform into each other. Moreover, HMG lexicons in the sense of \Def{def:hmg} could also be trained from large text corpora, e.g., in order to explain gradience effects. Then one has to check how subsymbolic harmony would be related to soft-constraint harmony obtained from corpus studies.

The requirements of our theory for tensor product constructions to be faithful representations of minimalist processing lead to extremely high-dimen\-sional embedding spaces. These spaces contain extremely few symbolically meaningful states. Therefore, numerical application on common workstations is only feasible by using compressed and thus non-faithful representations. Yet, non-faithful representations are also interesting for more principal reasons, as they allow for memory capacity constraints, e.g. by means of graceful saturation in neural network models \citep{SmolenskyLegendre06a, Smolensky06}. Several possible compression techniques have been suggested in the literature, e.g. contraction (i.e. outtraceing), circular convolution, holographic reduced representations, or geometric algebra \citep{CoeckeSadrzadehClark11, AertsCzachorMoor09, Plate03, SmolenskyLegendre06a, Smolensky06, GrabenPotthast09a}. It would therefore be necessary to generalize our current theory to compressed representations, including an assessment of the entailed representation errors. We leave this issue for future work.

Another important aspect of our work concerns the relationship between minimalist grammar and compositional semantics. On the one hand, it is straightforward to include semantic features into minimalist lexicons, e.g. as type-logical expressions \citep{NiyogiBerwick05}. On the other hand, this is somewhat redundant because the very same information is already encoded in the minimalist features \citep{Kobele06}. Vector space semantics appears as a very powerful tool for combining corpus-driven latent semantic analysis \citep{CederbergWiddows03} with compositional semantics based on compressed tensor product representations \citep{Blutner09, Aerts09, CoeckeSadrzadehClark11}. In our geometric representation theory, syntactic roles and thereby also semantic functions are encoded by node addresses in high-dimensional tensor products of role vectors for tree positions. Therefore, one should seek for appropriate unbinding maps that could be combined with their semantic counterparts \citep{CoeckeSadrzadehClark11}. Also this promising enterprise is left for future work.

% ------------------------------------- Section -----------------------------------
\section*{Acknowledgements}

This research was supported by a DFG Heisenberg grant awarded to PbG (GR 3711/1-1). We thank Peter Baumann, Reinhard Blutner,  Hans-Martin G\"artner and two referees for constructive suggestions.

% ------------------------------------- References -----------------------------------

% \bibliography{PbG}

\end{document}